\documentclass{article}
\usepackage[final]{corl_2017}

\usepackage[english]{babel}

\usepackage{graphicx}
\usepackage{animate}
\usepackage{epsfig} 				% for postscript graphics files
\usepackage{epstopdf}

\usepackage{listings}
\usepackage{color}
\usepackage{nameref}
\usepackage{hyperref}
\usepackage[colorinlistoftodos]{todonotes}
\usepackage{amsmath}	 			% assumes amsmath package installed
\usepackage{amssymb}  				% assumes amsmath package installed

\usepackage{dsfont}			%for the symbol 'Real Numbers'
\usepackage{mathtools}

\usepackage{epigraph}
\usepackage{lscape}
\usepackage[]{nomencl}				% nomenclatures
\usepackage{algorithm}
\usepackage{algorithmic}
\usepackage{multicol}
\usepackage{multirow}
\usepackage{etoolbox}

\usepackage{caption}
\usepackage{subcaption}

\usepackage{wrapfig}
\usepackage{multimedia}

\usepackage{units}
\usepackage{flushend}

\usepackage{floatflt}

\usepackage{url}
\usepackage[absolute,overlay]{textpos}

\usepackage{bibentry}

\usepackage{tikz}
\usepgflibrary{arrows}	% for more options on arrows

\usepackage{float}
\usepackage[utf8]{inputenc}
\usepackage[english]{babel}
\usepackage{graphics}
\usepackage{animate}

\usepackage{listings}

\usepackage[nostamp]{draftwatermark}

\usepackage{extarrows}		%for text above eq

\allowdisplaybreaks[1]

\newcommand{\playvideo}[1]{\href{run:#1}{\includegraphics[scale=0.12]{\RCPath fig/empty}}}

\usepackage{lipsum}
\usepackage{enumitem}
\usepackage{framed}

\newcommand{\ie}[0]{i.e.}
\newcommand{\eg}[0]{e.g.}

\newcommand{\email}[1]{\href{mailto:#1}{\nolinkurl{#1}}}

\newcommand{\link}[1]{\colora{\url{#1}}}

\renewcommand{\sec}[1]{Section~\ref{#1}}
\newcommand{\fig}[1]{Figure~\ref{#1}}

\newcommand{\tab}[1]{Table~\ref{#1}}

\definecolor{matlab1}{rgb}{0,0,1}
\definecolor{matlab2}{rgb}{0,0.5,0}
\definecolor{matlab3}{rgb}{1,0,0}
\definecolor{matlab4}{rgb}{0,0.75,0.75}
\definecolor{matlab5}{rgb}{0.75,0,0.75}
\definecolor{matlab6}{rgb}{0.75,0.75,0}
\definecolor{matlab7}{rgb}{0.25,0.25,0.25}

\definecolor{darkgreen}{rgb}{0,0.5,0}		%Olivegreen?
\definecolor{purple}{rgb}{0.75,0,0.75}
\definecolor{pink}{rgb}{1,0.4,0.6}

\newcommand{\capitalize}[1]{\expandafter\MakeUppercase\expandafter{#1}}

\newcommand{\colora}[1]{{\usebeamercolor[fg]{framesubtitle}#1}}

\makeatletter
\newcommand*{\compress}{\@minipagetrue}
\makeatother

\renewcommand{\vec}[1]{\boldsymbol{#1}}				% Vector
\newcommand{\inputSca}[0]{x} % x
\newcommand{\inputVec}[0]{\vec{\inputSca}}

\newcommand{\outputSca}[0]{y}

\newcommand{\regressionNo}[0]{f}
\newcommand{\regression}[1]{\regressionNo \left( #1 \right)}

\newcommand{\parameter}[0]{\theta} 				% Parameter
\newcommand{\parameters}[0]{\vec{\parameter}} 			% Parameters
\newcommand{\q}{\vec{q}}					% Position
\ifdef{\dq}{\renewcommand{\dq}{\dot{\q}}}{\newcommand{\dq}{\dot{\q}}}
\usepackage{array}
\makeatletter
\newcommand{\thickhline}{%
    \noalign {\ifnum 0=`}\fi \hrule height 1pt
    \futurelet \reserved@a \@xhline
}
\newcolumntype{"}{@{\hskip\tabcolsep\vrule width 1pt\hskip\tabcolsep}}
\makeatother

\newcommand{\website}[0]{\url{https://lasr.org/research/feeling-of-success}}

\newcommand\blfootnote[1]{%
  \begingroup
  \renewcommand\thefootnote{}\footnote{#1}%
  \addtocounter{footnote}{-1}%
  \endgroup
}

\newcommand{\approxndatagrasping}[0]{{9,000}}
\newcommand{\exactndatagrasping}[0]{{9269}}
\newcommand{\gelsight}[0]{{GelSight}}
\newcommand{\forcegripper}[0]{F}
\newcommand{\anglegripper}[0]{\phi}

\newcommand{\nCollectedObj}[0]{106}

\newcommand{\nGraspingObj}[0]{12}

\newcommand{\successVisTac}[0]{94}
\newcommand{\successVis}[0]{80}

\title{The Feeling of Success:\\Does Touch Sensing Help Predict Grasp Outcomes?}

\author{
  Roberto Calandra\footnote{}\\
  U.C. Berkeley\\
  \And
  Andrew Owens\\
    U.C. Berkeley\\
  \And
  Manu Upadhyaya \\
  U.C. Berkeley\\
  \And
  Wenzhen Yuan\\
  MIT\\
    \And
  Justin  Lin\\
  U.C. Berkeley\\
  \And
  Edward H. Adelson\\
  MIT\\
  \And
  Sergey Levine\\
  U.C. Berkeley\\
  }

\begin{document}
\maketitle

%===============================================================================

\vspace{-0.8cm}
\begin{abstract}
  A successful grasp requires careful balancing of the contact forces. 
Deducing if a particular grasp will be successful from indirect measurements, such as vision, is quite challenging, and direct sensing of contacts through touch sensing provides an appealing avenue toward more successful and consistent robotic grasping. 
However, to fully evaluate the value of touch sensing for grasp outcome prediction, we must understand how touch sensing can influence outcome prediction accuracy when combined with other modalities. 
Doing so using conventional model-based techniques is exceptionally difficult. 
In this work, we investigate the question of whether touch sensing aids in predicting grasp outcomes within a multimodal sensing framework that combines vision and touch. 
To that end, we collected more than \approxndatagrasping{} grasping trials using a two-finger gripper equipped with \gelsight{} high-resolution tactile sensors on each finger, and evaluated visuo-tactile deep neural network models to directly predict grasp outcomes from either modality individually, and from both modalities together.
Experimental results indicate that incorporating tactile readings substantially improve grasping performance.
Code and dataset is available online at \website{}
\end{abstract}

\keywords{Robot Learning, Grasping, Tactile Sensors, Multi-modal Sensing} 

%===============================================================================

\section{Introduction}

\begin{wrapfigure}{r}{0.30\textwidth}
\vspace{-10pt}
	\centering
	\includegraphics[width=0.90\linewidth]{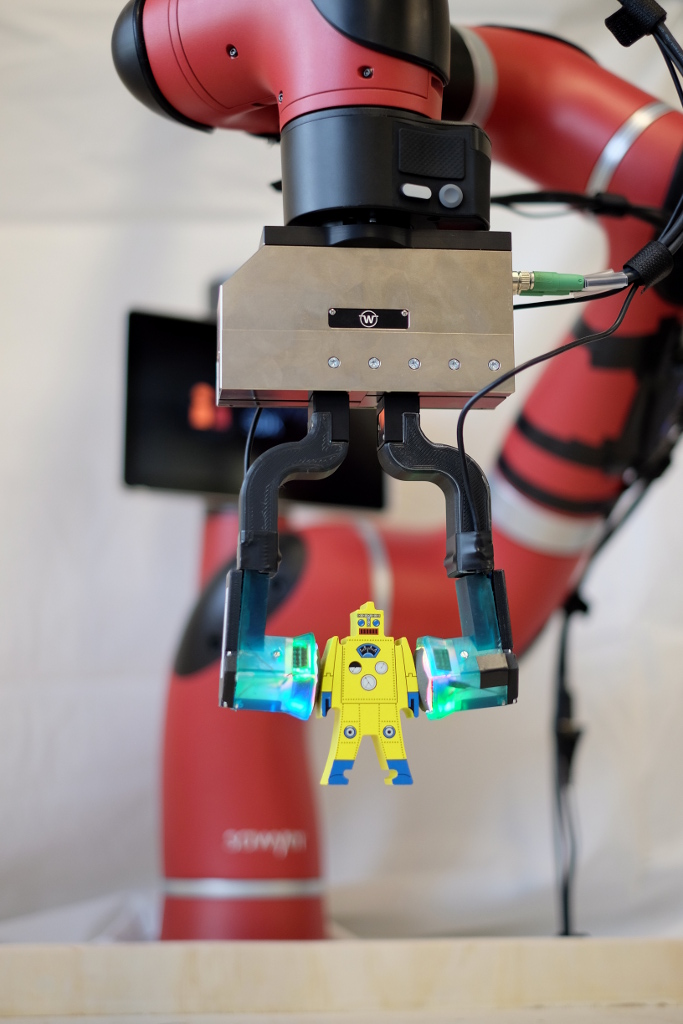}
    \caption{The use of tactile sensors can greatly improve robot grasping capabilities. In our experiments, we used two \gelsight{} sensors mounted on a parallel jaw gripper.}
	\label{fig:setting}
	\vspace{-40pt}
\end{wrapfigure}
Humans\blfootnote{$^*$ Corresponding author: \email{rcalandra@lasr.org}} make extensive use of multi-modal perception when grasping, including visual and tactile sensing~\cite{Johansson2009}.
While vision allows fast localization of objects, touch provides accurate perception of compliance and contact force once contact is established, even when the grasp itself is hard to see~\citep{Howe1993}.
In manipulation, the use of tactile sensors has been demonstrated for detecting and compensating for slip~\citep{Bicchi1988,Romano2011,Veiga2015}, and for grasping fragile objects~\citep{Stachowsky2016}.
Nonetheless, adoption of tactile sensors has been slow, due to hardware limitations of the technologies employed (\eg, sensitivity, cost) and, more importantly, due to the challenges associated with integrating tactile sensors into standard control schemes.
Tactile readings are typically difficult to model, and and small errors in measurement and calibration can substantially reduce the performance of even the best analytic models.

These modeling issues echo the challenges that, until recently, have also plagued vision-based sensing. 
Recently, a number of works have proposed vision-based grasping methods that rely on end-to-end training, rather than analytic modeling, to directly predict grasp locations or grasp outcomes based on camera images~\citep{Lenz2015,Pinto2016,Levine2016}.
However, the use of learning to process tactile readings for grasping, particularly with end-to-end models such as deep neural networks, has not been studied extensively, despite the importance of this modality.

% Contribution
In this paper, we aim to answer the question of whether integrating touch sensing aids in predicting grasp outcomes. 
However, the answer to this question is complicated by the inherent difficulty of integrating vision and touch into a multi-modal sensing framework. 
We propose to employ an end-to-end learning approach for predicting grasp outcome, which allows us to evaluate the relative importance of each modality, as well as the combined multi-modal framework, in a series of controlled experiments.
For touch sensing, we employ the \gelsight{}~\citep{Dong2017} tactile sensors (see \fig{fig:setting}), which provide high-resolution images of the deformation caused by contacts with graspable objects (see examples in \fig{fig:tactile_data}).
We train deep neural networks for vision-based, touch-based, and combined vision and touch-based grasp outcome prediction.
A robot can then employ these models for grasping by attempting various grasp locations, and choosing the one for which the model predicts the highest probability of success.
To our knowledge, this work is the first to present an end-to-end learned system for robotic grasping that combines rich visual and tactile sensing, and provides a controlled evaluation of the benefits of touch sensing for grasp performance.
Our method is substantially simpler than manually designed analytic grasping metrics (\eg,~\citep{Rosales2011}), and the experimental results demonstrate that incorporating tactile sensing improves overall grasping performance substantially.

%===============================================================================

\section{Related Work}
\label{sec:related}

\begin{figure}[t]
	\centering
	\includegraphics[height=2cm]{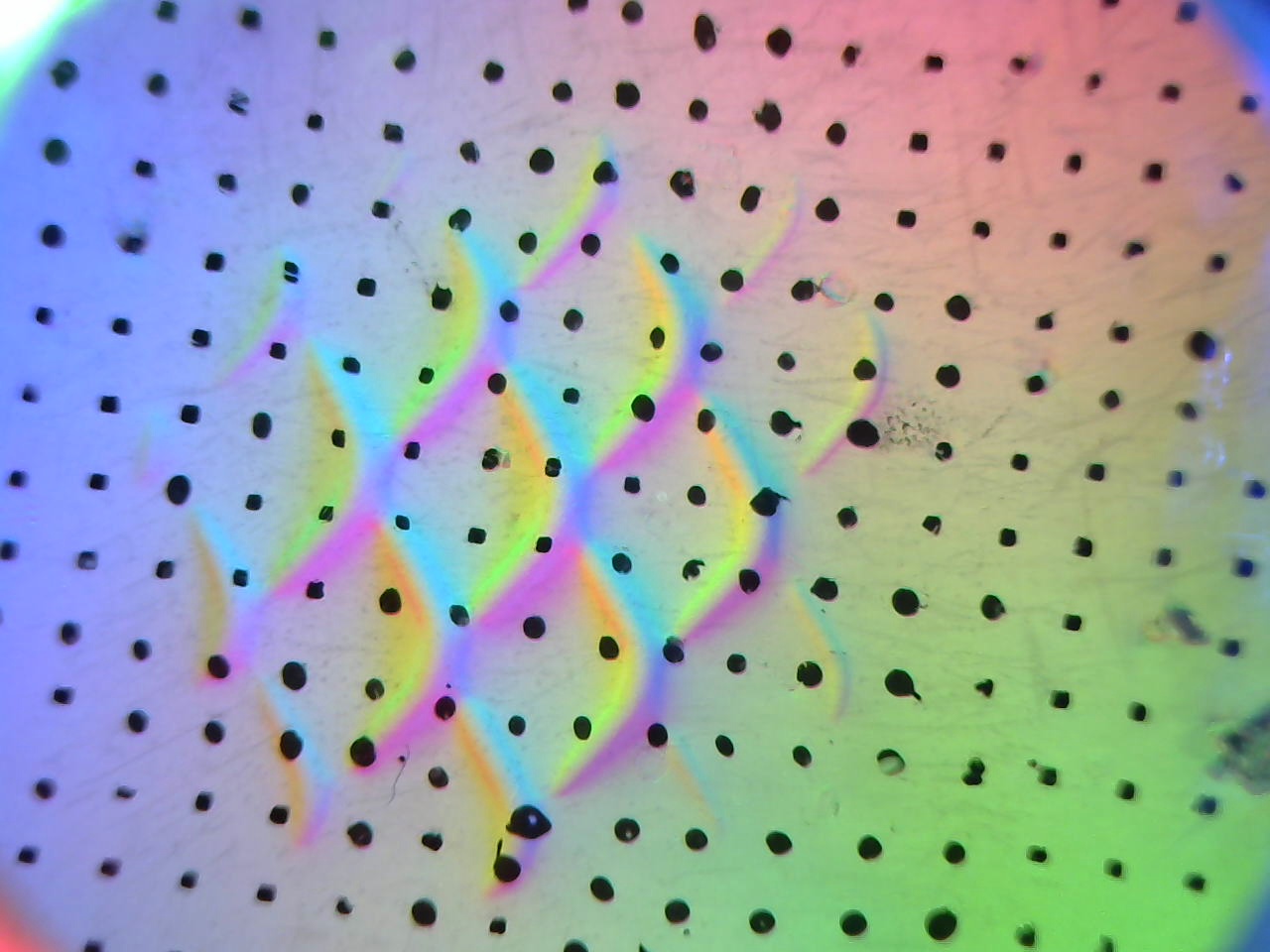} \hfill
	\includegraphics[height=2cm]{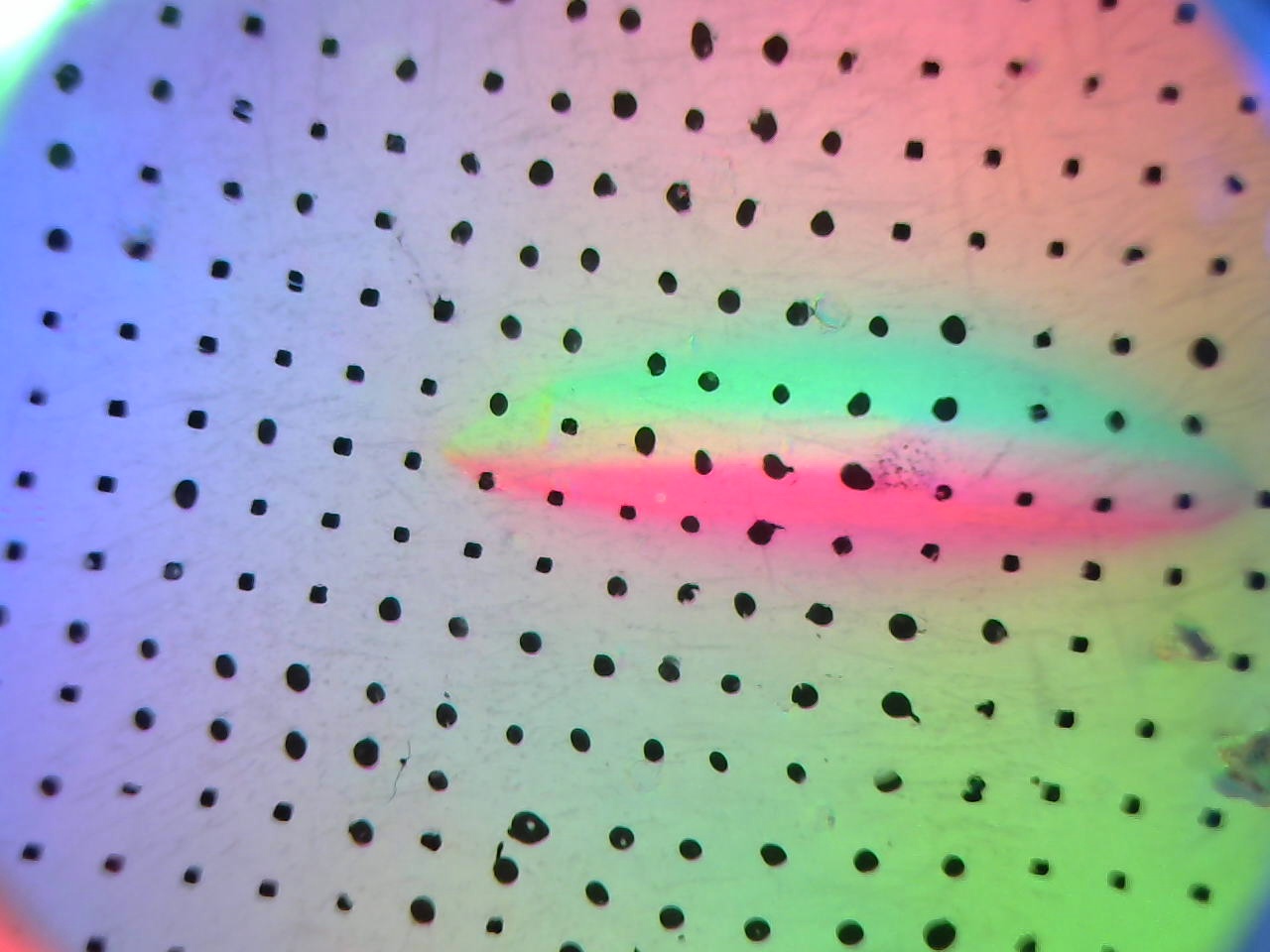} \hfill
	\includegraphics[height=2cm]{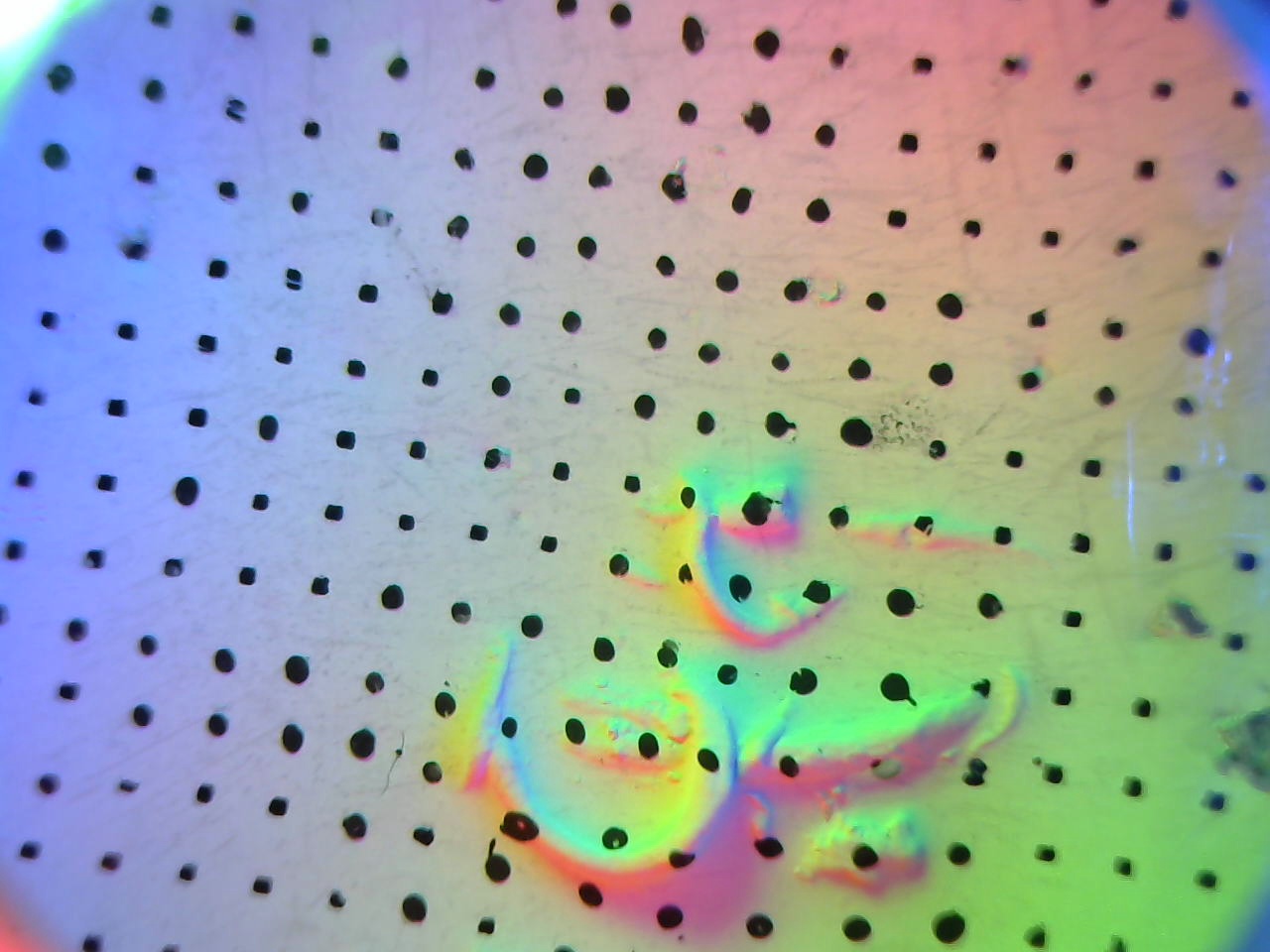} \hfill
	\includegraphics[height=2cm]{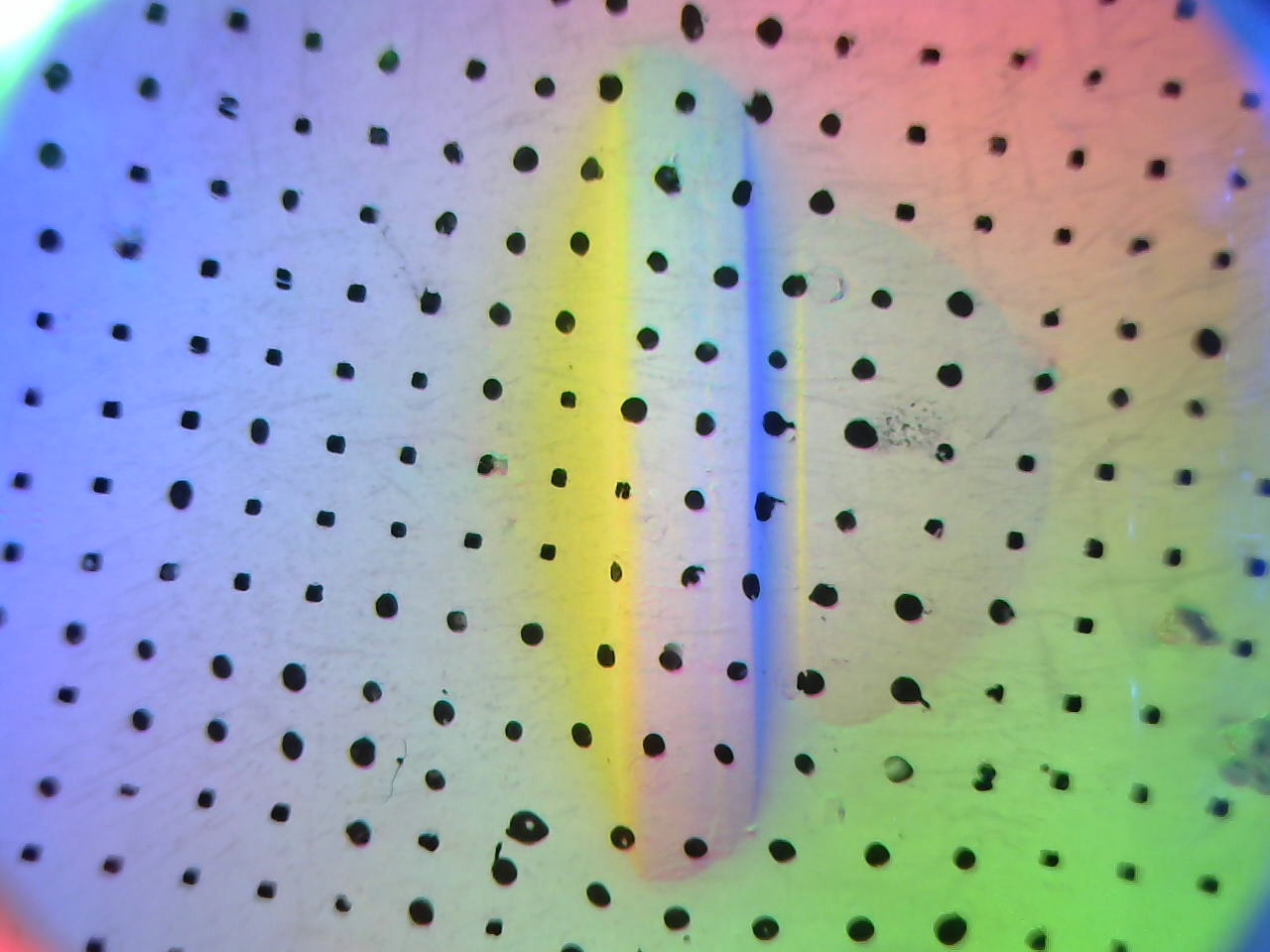} \hfill
	\includegraphics[height=2cm]{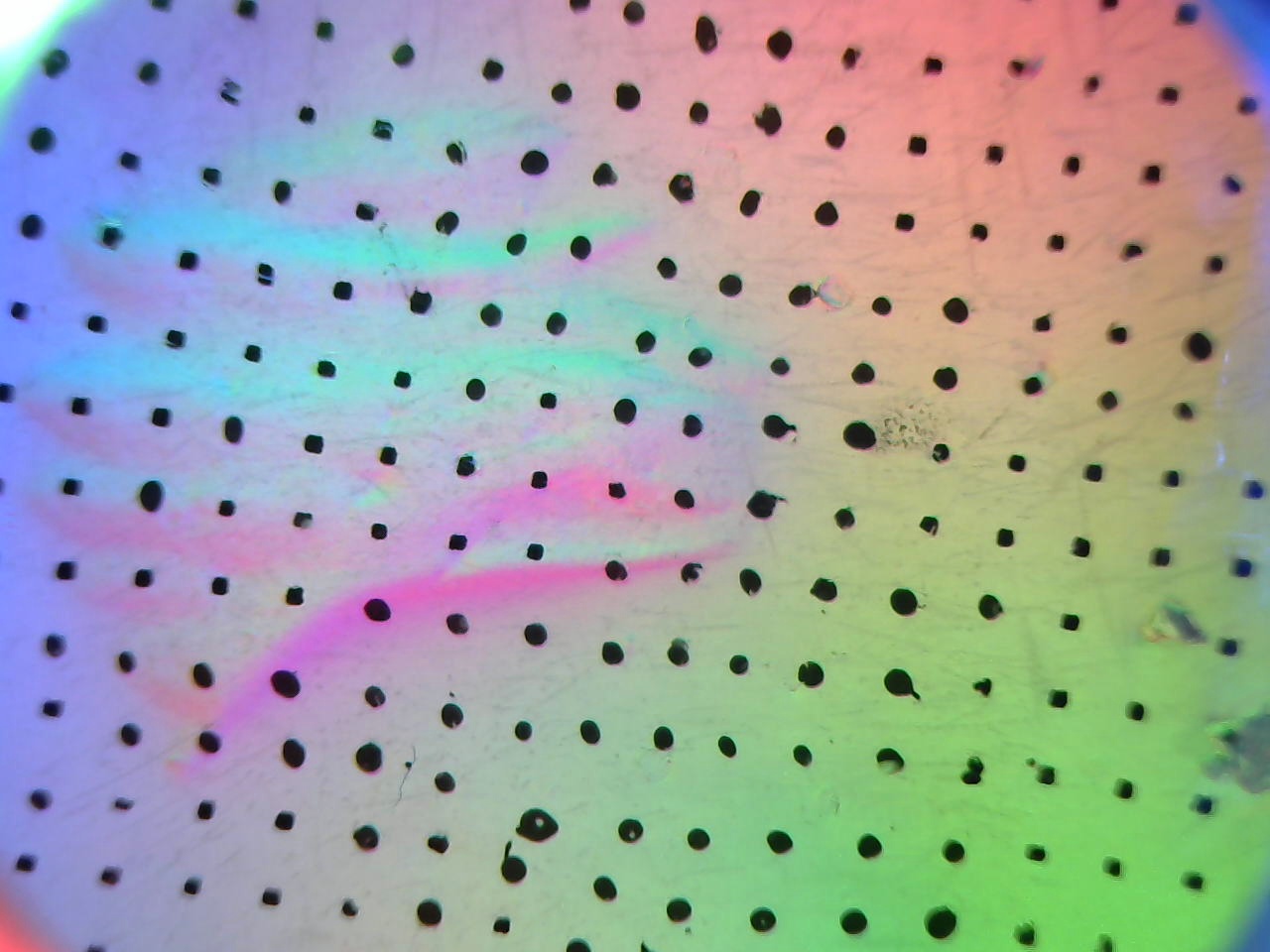}
	\\ \vspace{3pt}
	\includegraphics[height=2cm]{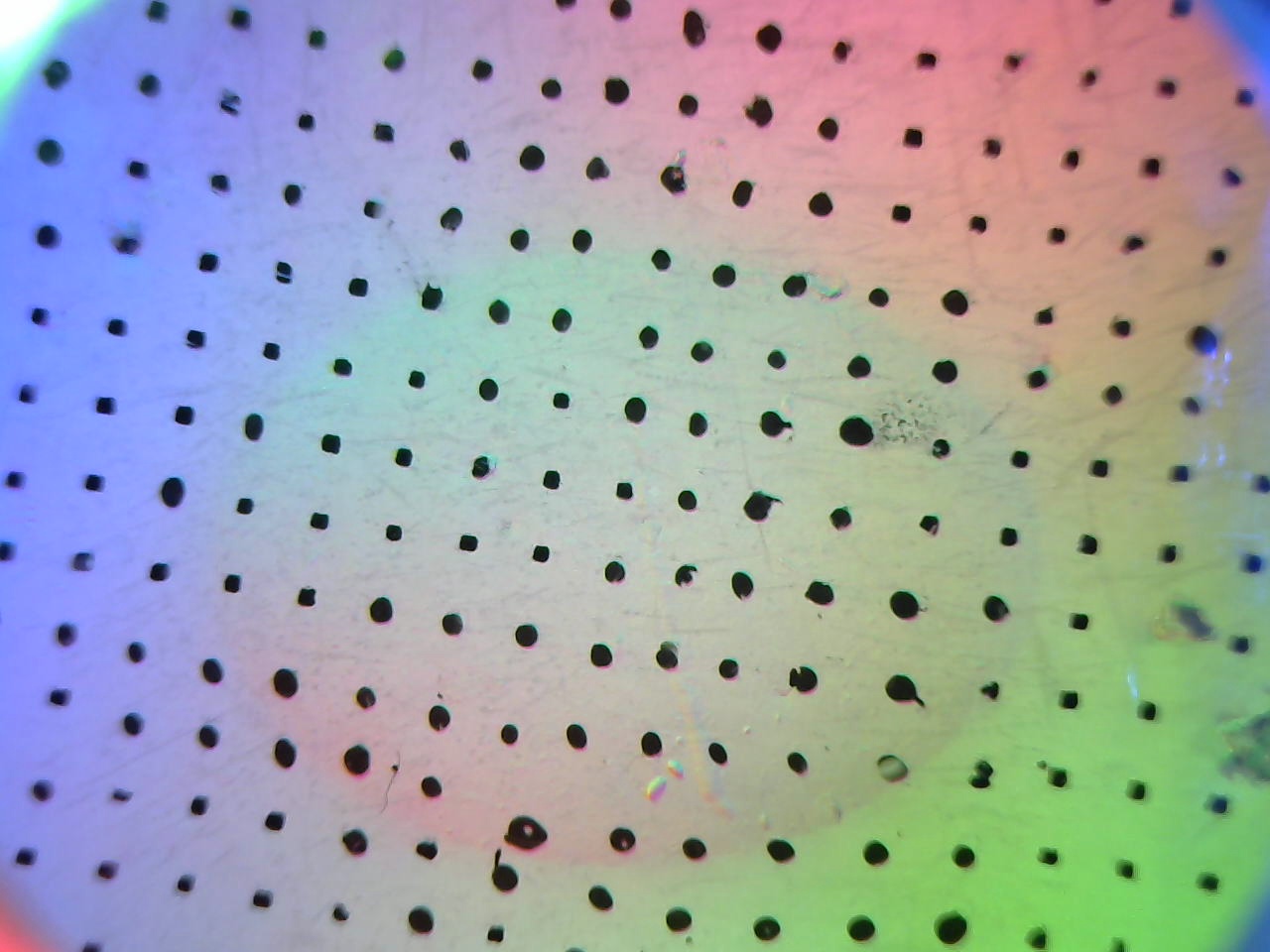} \hfill
	\includegraphics[height=2cm]{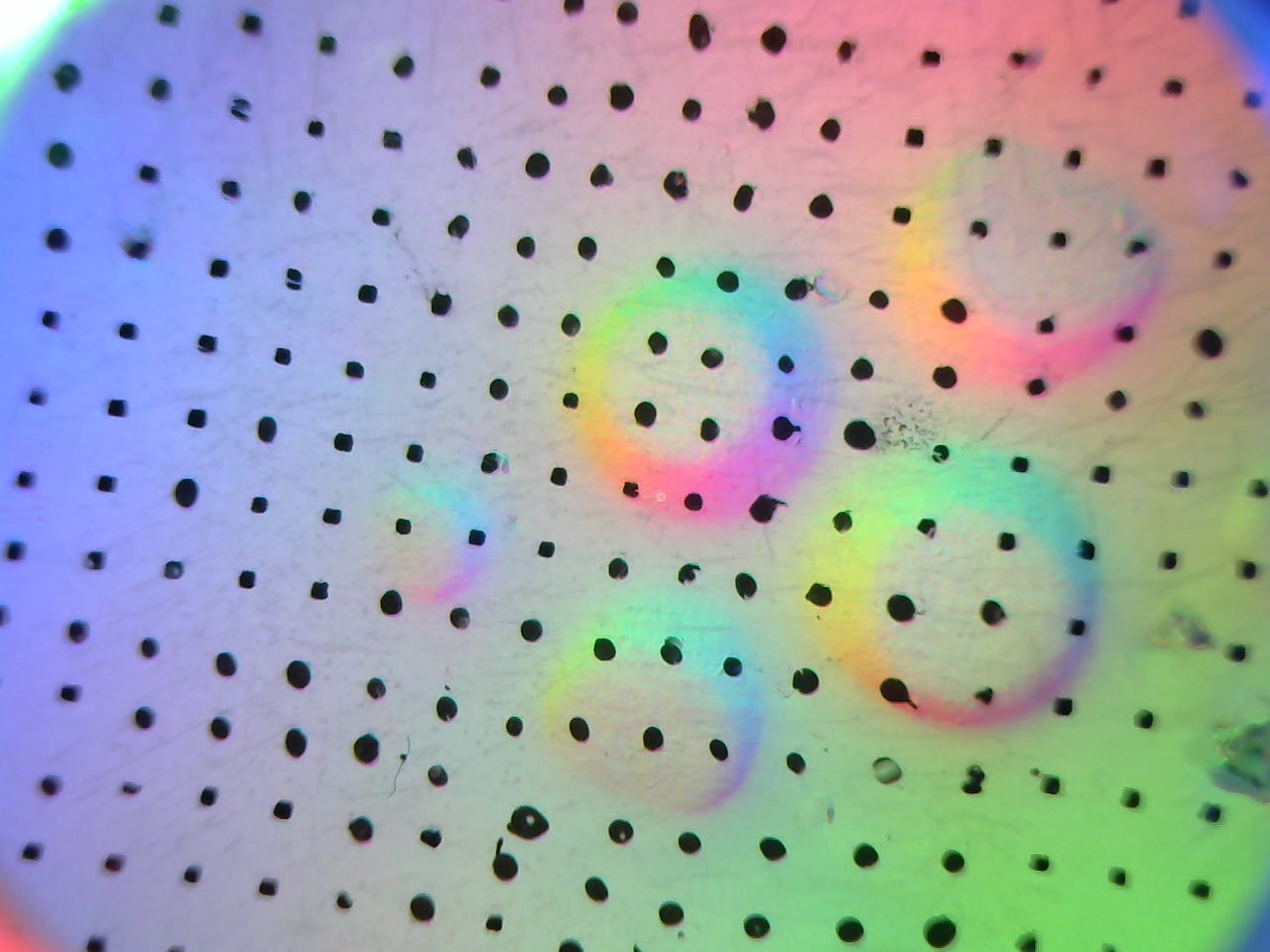} \hfill
	\includegraphics[height=2cm]{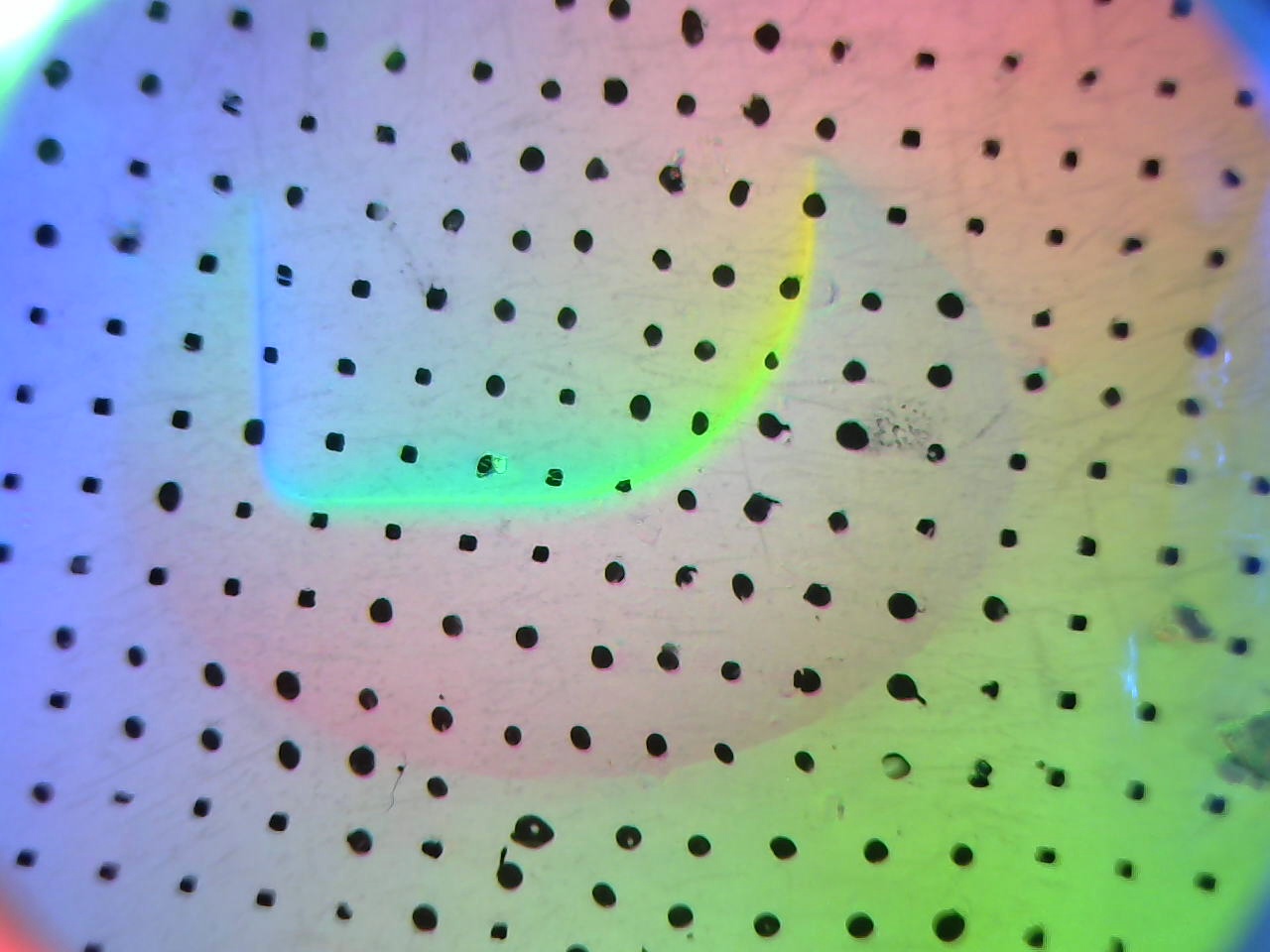} \hfill
	\includegraphics[height=2cm]{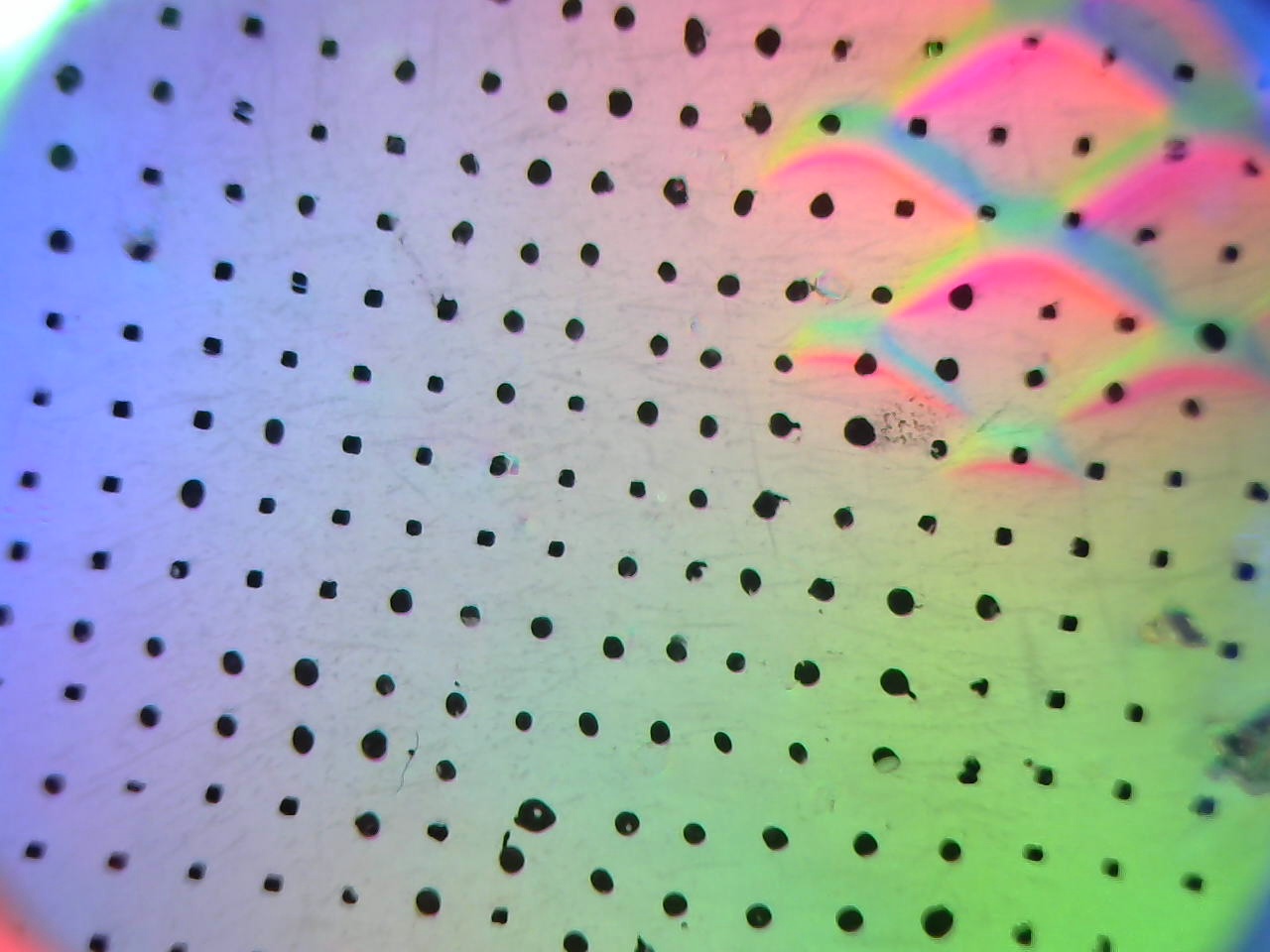} \hfill
	\includegraphics[height=2cm]{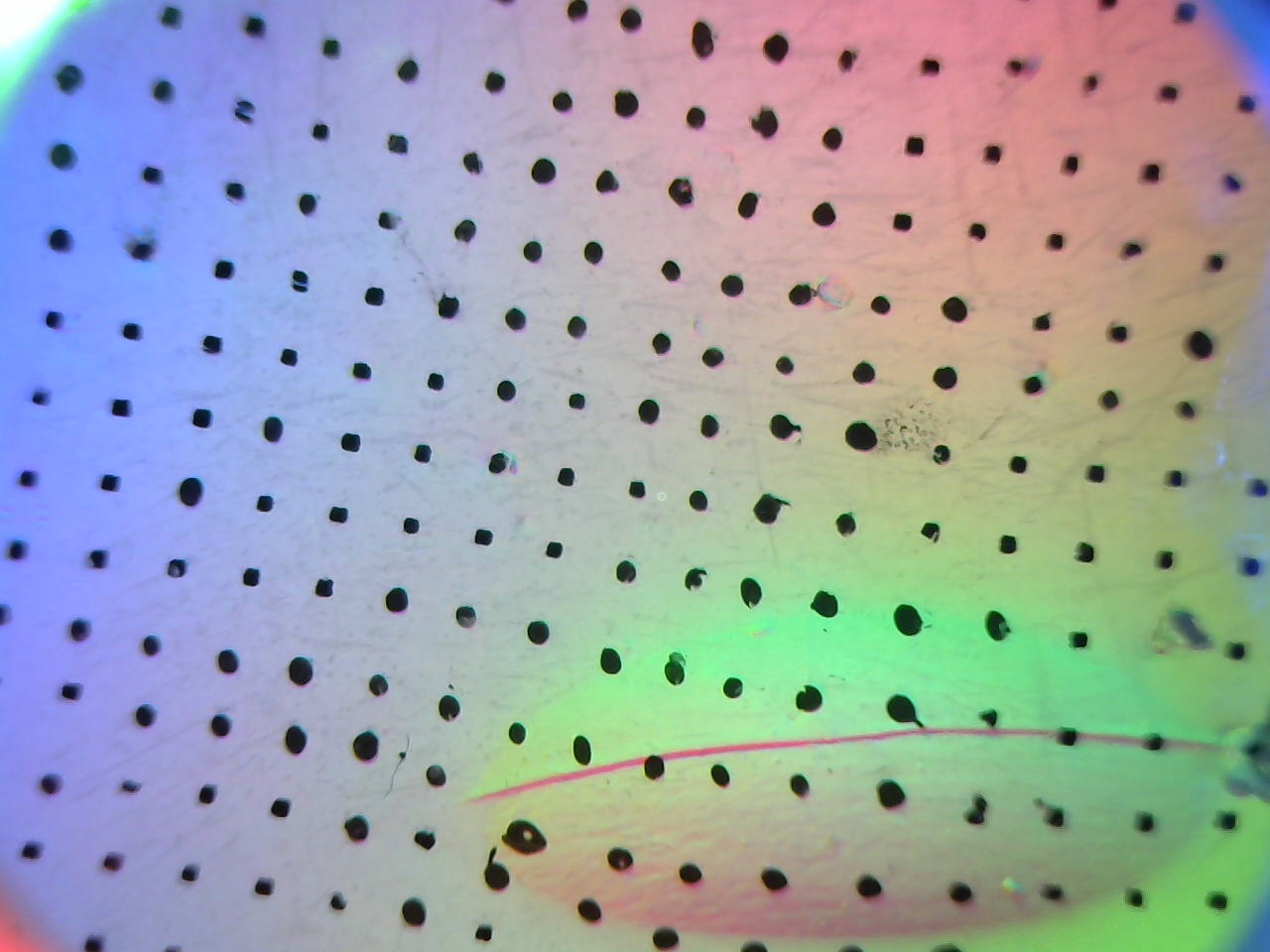}
	\caption{Examples of raw tactile data collected by the \gelsight{} for different training objects.}
	\label{fig:tactile_data}
\end{figure}

\vspace{-0.1in}
\paragraph{Learning to grasp.}
	A significant body of work in robotics has studied analytic grasping models, which use models of object geometry, environments, and robot grippers, and which typically make use of a manually defined grasping metric~\citep{Shimoga1996,Goldfeder2011,Rodriguez2012}. 
	While these methods provide considerable insight into the physical interactions in grasping, the relationship between grasp metrics and real-world outcomes is vulnerable to model misspecification and unmodeled effects. 
	As an alternative, data-driven approaches have sought to predict grasp outcomes from human supervision~\citep{Kamon1996,Saxena2008,Lenz2015}, simulation~\citep{Kappler2015,Johns2016,Mahler2016}, or autonomous robotic data collection~\citep{Pinto2016,Levine2016}, typically using visual or depth observations. 
	However, these methods have not been applied to tactile sensing, and therefore have limited ability to reason about contact forces, pressures, and compliance. 
	As a substitute for this capability, some works have used wrist-mounted depth cameras to obtain a better estimate of local geometry~\citep{Viereck2017}.
	\citet{Gualtieri2016} detected grasping poses from depth data, using convolutional neural network trained on simulated data. 
	In this work, we use over-the-shoulder cameras, and show that we can obtain substantial improvement from incorporating rich tactile sensing.
	For a more detailed survey on learning to grasp we refer the readers to~\citet{Bohg2014}.

\vspace{-0.1in}
\paragraph{Tactile sensors in grasping.}
	A range of touch sensors have been proposed in the literature~\citep{Yousef2011}, and they have been employed in a variety of ways to aid robotic grasping.
	For example, \citet{Bekiroglu2011} and \citet{Schill2012} used tactile sensors to estimate grasp stability, and \citet{Li2014a} proposed incorporating tactile readings into dynamics models of objects for a dexterous hand.
	Works such as \citep{Bicchi1988,Romano2011,Veiga2015,Stachowsky2016} extract features from tactile signals to detect slip, so as to apply a proper grasping force. 
	Researchers have also proposed robotic systems that integrate visual and tactile information for grasping using model-based methods~\citep{Allen1999,Bekiroglu2012,Jara2014}, which improved the grasping performance over single-modality inputs.
	However, to the best of our knowledge, our work is the first to propose end-to-end training of models that process rich visual and tactile inputs to predict task outcomes, which in our case corresponds to predicting if a grasp will succeed or fail, and is also the first to provide a controlled evaluation of whether incorporating touch actually improves grasp success and outcome prediction within a learned visuo-tactile system.
	% There are also works on multi-modal learning using both visual and tactile inputs for object property estimation~\citep{Kroemer2011,Gao2016}. 

\vspace{-0.1in}
\paragraph{The GelSight sensor.}
	% Describe the gelsights 
	The \gelsight{} sensor is an optical tactile sensor that measures high-resolution topography of the contact surface~\citep{Johnson2009,Dong2017}. 
	The surface of the sensor is a soft elastomer painted with a reflective membrane, which deforms to the shape of the object upon contact. 
	Underneath this elastomer is an ordinary webcam that views the deformed gel. 
	The gel is illuminated by colored LEDs, which light the gel from different directions.
	\citet{Dong2017} showed that for the grasping tasks, the \gelsight{} signal can predict or detect slip from 3 different kinds of information: movement of the object texture, loss of contact area, and the stretching of the sensor surface.
	One important advantage of the \gelsight{} sensor is that the sensory data consists of a standard 2D image, allowing for standard convolutional neural network architectures designed for visual sensing to be used to process the sensor's readings.
	Prior work on material estimation with the \gelsight{}~\citep{Yuan2017,Yuan2017a} successfully applied convolutional neural networks that were pretrained from natural image data.
	Examples of raw tactile data from the \gelsight{} are shown in \fig{fig:tactile_data}.

%===============================================================================

\section{Predicting Successful Grasps from Vision and Touch}
\label{sec:model}

\begin{figure}[t]
	\centering
	\includegraphics[width=0.98\linewidth]{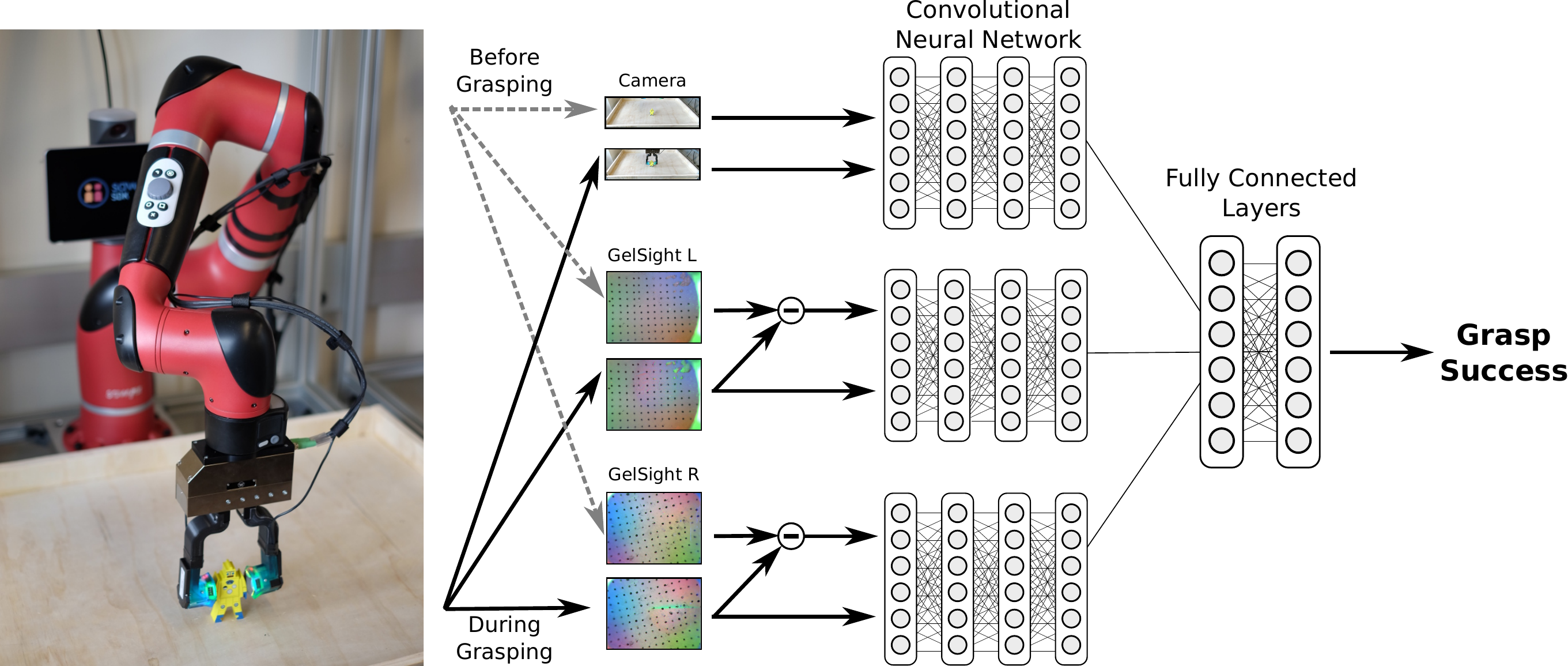}
	\caption{Diagram of our visual-tactile multi-modal model. At grasping time (before attempting to lift the object), the RGB images from the front camera and the \gelsight{} sensors images are fed to a deep neural network which predict whether the grasping will be successful or not. In the network, the data from each of the sensors is first passed into a convolutional neural network, and the resulting features are concatenated as the input into a fully-connected network.}
	\label{fig:diagram}
\end{figure}
Consider the placement of the robot's gripper in \fig{fig:diagram}. 
If the robot attempts to grasp the object from this position, will it be successful? 
Here, the sensory modalities give different -- and possibly complementary -- information about the prospects for a successful grasp.  
For example, we can tell from the camera image that the gripper is in a favorable position -- near the object's center of mass -- and we can also tell from the tactile information that the object is rigid, with many bumps and ridges, and therefore is unlikely to slip. 
To study how much tactile sensing helps us predict grasp outcomes, we train a neural network to predict whether a robot's grasp will be successful using a combination of tactile and visual cues. 
This network computes $\outputSca = \regression{\inputVec}$, where $\outputSca$ is the probability of a successful grasp, and $\inputVec$ contains a set of images from multiple modalities: $\inputVec = (I_{\text{RGB}}, I_{\text{GelsightL}}, I_{\text{GelsightR}})$, where $I_{\text{RGB}}$ represents an RGB image from the frontal camera, and $I_{\text{GelsightL}}$ and $I_{\text{GelsightR}}$ are the RGB images recorded by the two fingertip \gelsight{} sensors.  
In our experiments (see \sec{sec:result}), we compared against different combinations of inputs, such as only vision, only tactile input or using a depth map in place of the RGB image. 

Since all of the inputs to the model -- including the tactile input -- are images, we represent $\regressionNo$ as a convolutional neural network.  
Following other multi-modal learning work~\cite{Ngiam2011}, we fuse the different modalities at a late stage in the model.  
As illustrated in \fig{fig:diagram}, the images are first passed through a standard convolutional network, which in our case uses the ResNet-50 architecture \cite{he2016deep} (fine-tuned during the training, as detailed below).
The results of these independent convolutions are then concatenated and fed into a two-layer, fully-connected network. 
For both the visual and tactile inputs, we make use of temporal information.  
For the RGB images, we simply supply the network with two images: an image $I_{T_a}$ taken before the grasp (where the object is unoccluded), and one $I_{T_b}$ at the moment of the grasp (at which point the gripper has been placed on the object).  
The features from both networks -- specifically, the spatially-pooled features from the penultimate layer of each ResNet model -- are then concatenated together. 
For the \gelsight{} model, we exploit the fact that deformations in the gel can be expressed as temporal derivatives, and apply the network to the temporal difference $I_{T_b} - I_{T_a}$.

% Training network
\paragraph{Training the network} We initialize the weights of both the visual and tactile CNNs using a model pre-trained on ImageNet~\cite{Deng2009}, and we share parameters between networks of the same modality (e.g., both GelSight sensors use the same network weights). We train the network or 20 epochs using the Adam optimizer~\cite{Kingma2014}, starting with a learning rate of $10^{-4}$ (which we decrease by an order of magnitude halfway through the training process).

During training, we apply data augmentation to the input data. 
For the RGB images, we first crop the images with a bounding box containing the table that holds the objects(padded vertically to provide a view of the robot's gripper).  
Then, following standard practice in object recognition, we resize these images to be $256 \times 256$, and randomly sample $224 \times 224$ crops from them.  We also randomly flip the images in the horizontal direction.  The \gelsight{} images lack the stationarity properties of natural images, due to the fixed locations of the lights and the borders of the gel.  
However, we still apply these same data augmentation techniques to prevent the algorithm from overfitting to the appearance of a particular sensor's gel.

%===============================================================================

\section{Grasping with Vision and Touch}
\label{sec:optimization}

	In this section, we detail how the model learned in \sec{sec:model} can be used for selecting grasping configurations.
Since the model predicts the grasp outcome based on visual and tactile readings, we must close the fingers around an object before we can evaluate the model's prediction. 
To that end, in our experimental comparison we perform randomly chosen gripper closures in the vicinity of the object, evaluate the prediction of the model on each one, and accept the gripper pose for which the model predicts a sufficiently high probability of a successful outcome.
Specifically, we randomly vary the grasp parameters $\parameters = [\text{EE}_x,\text{EE}_y,\text{EE}_z,\anglegripper, \forcegripper]$, where $[\text{EE}_x,\text{EE}_y,\text{EE}_z]$ are the end-effector x-y-z coordinates, $\anglegripper\in[0,\pi]$ is the angle of the gripper, and $\forcegripper$ is the force applied by the gripper.
At each iteration, we randomly select a set of parameters, move the gripper to the desired configuration, close the gripper, and then evaluate the success rate predicted by the model (e.g., using the visual and tactile data described in \sec{sec:model}) for the measured visual and tactile readings.
If the success rate is above a pre-defined threshold ($0.9$ in our experiments), the grasp is considered potentially successful, and the object is lifted to observe the outcome of the grasp.
Otherwise, a new random set of parameters is selected and the exploration continues.
While in theory this approach might never stop, in practice, in our experiments most trials would last less then a couple of minutes. 
Overall, this scheme can be intuitively summarized as having two separate mechanism, one which proposes grasps (\ie, random search), and one that rejects them (\ie, the model); The process is over once a proposal is accepted. 
In future work, this process might be accelerated by incorporating an optimization procedure into the grasp selection, instead of proposing grasps at random. 
However, for the purposes of evaluating the practical differences between purely visual and visuo-tactile grasping, we found that this simple approach was sufficient.

%===============================================================================
	
\section{Experimental Setting \& Data Collection}
\label{sec:setting}

\begin{figure}[t]
	\centering
	\includegraphics[width=0.99\linewidth]{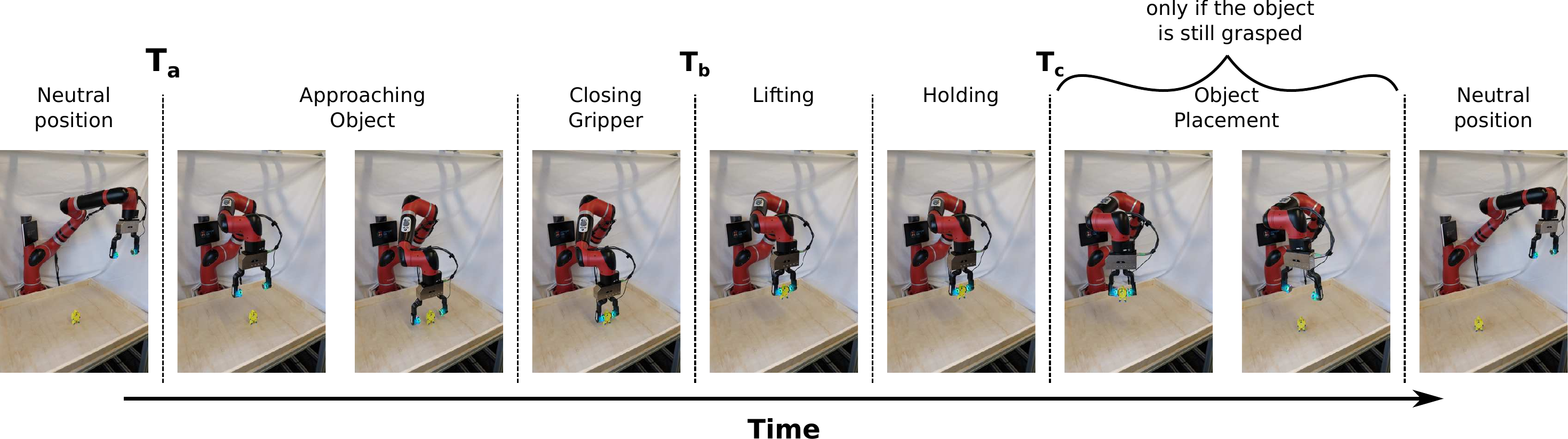}
	\caption{Chronology of a data collection trial, with the various grasping phases, and the three snapshot points $T_a, T_b, T_c$.}
	\label{fig:time}
	\vspace{-10pt}
\end{figure}
%

% Describe the hardware setting
In our experiments we used a setup, shown in \fig{fig:setting}, consisting of a 7-DoF Sawyer arm, a Weiss WSG-50 parallel gripper, and two \gelsight{} sensors, one for each finger.
The design of the \gelsight{} sensors we used in this project is introduced in~\citep{Dong2017}. 
The sensors provides raw-pixel measurements at a resolution of 1280x960 @30Hz over an area of 24x18 mm. 
Additionally, one Microsoft Kinect~2 sensor was mounted in front of the robot.
It should be noted that in our visuo-tactile model only the RGB image is being used, while the depth is being used in our hand-engineered data collection procedure (as detailed below), and as a comparison in \sec{sec:4:classifier}.

% Data collection
The data collection process was automated to allow for large scale continuous data collection.
In each trial, depth data from the Kinect was used to approximately identify the position of the object and fit a rough cylindrical proxy.
We then selected the grasp positions $[\text{EE}_x,\text{EE}_y]$ to be the center of the cylinder, plus a small random perturbation. 
The height $\text{EE}_z$ was set to a random height between the table and the top of the cylinder, and~$\phi$ was set to a random gripper orientation.
Moreover, we randomized the gripping force~$\forcegripper$ to collect a large variety of behaviors, from firm, stable grasps, to occasional slips, to overly gentle grasps that fail more often.
After moving to the chosen position and orientation and closing the gripper with the desired gripping force, the gripper would then attempt to lift the object and wait in the air for two seconds.
If the object was still in the gripper at the end of the two seconds, the robot would then put the object back at a randomized position, and a new trial would start.
During each trial we considered three snapshots: 
1) $T_a$ is the initial state of the system (with the arm in rest position, outside the view of the camera). 
2) $T_b$ measures the state when the gripper completed the closure of the fingers, but the object is still on the ground. 
3) Finally, $T_c$ is measured two seconds after the completed lift-off, to give time to the object to stabilize or eventually slip). 
Of these three snapshots, $T_a$ and $T_b$ are used as inputs to our models, while $T_c$ is used to label the data.
A visualization of the chronology of the data collection and of the three snapshots is shown in \fig{fig:time}.
Overall, each trial took $\sim60$ seconds of robot execution.

\begin{figure}[t]
	\centering		
	\includegraphics[height=2.25cm, width=1.5cm]{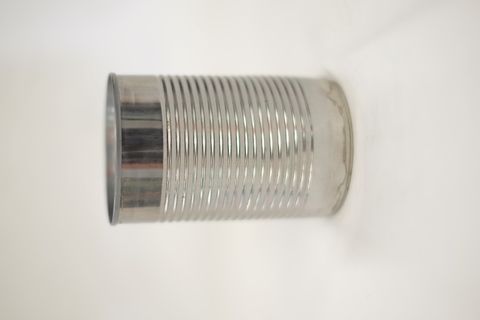} \hspace{1pt}
	\includegraphics[height=2.25cm, width=1.5cm]{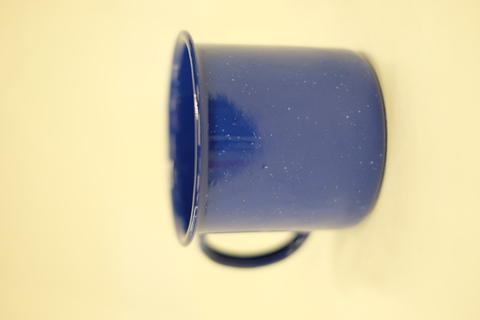} \hspace{1pt}
	\includegraphics[height=2.25cm, width=1.5cm]{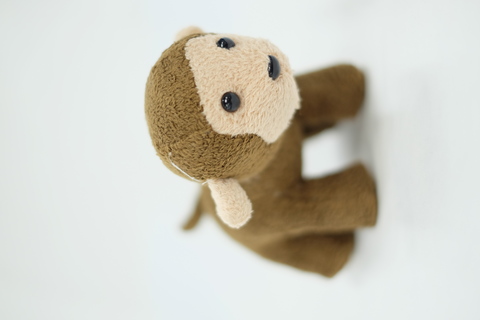} \hspace{1pt}
	\includegraphics[height=2.25cm, width=1.5cm]{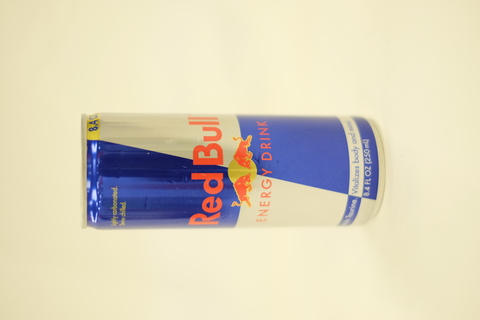} \hspace{1pt}
	\includegraphics[height=2.25cm, width=1.5cm]{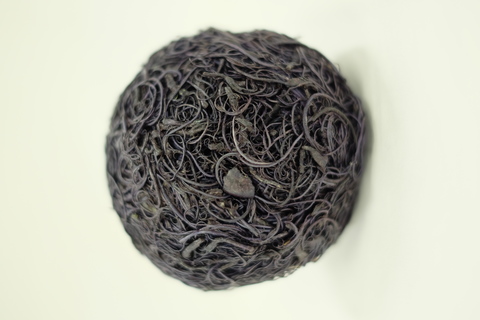} \hspace{1pt}
	\includegraphics[height=2.25cm, width=1.5cm]{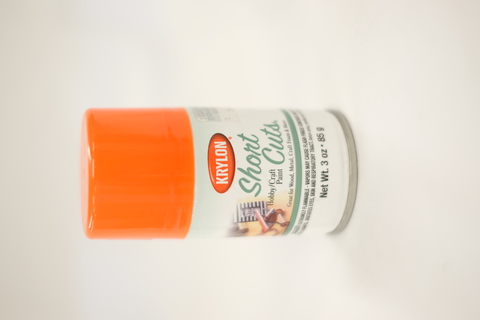} \hspace{1pt}
	\includegraphics[height=2.25cm, width=1.5cm]{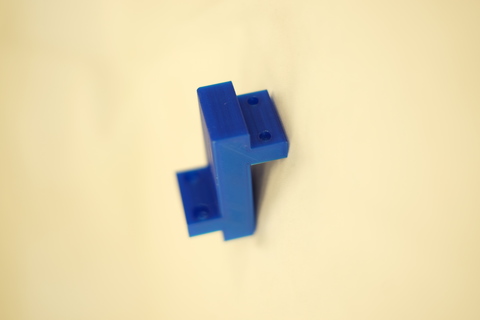} \hspace{1pt}
	\includegraphics[height=2.25cm, width=1.5cm]{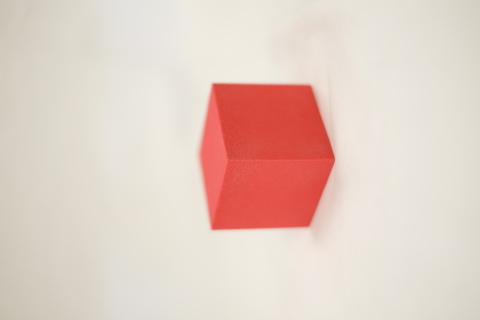}
	\\ \vspace{5pt}
	\includegraphics[height=2.25cm, width=1.5cm]{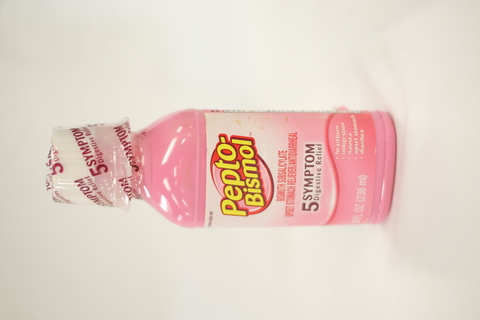} \hspace{1pt}
	\includegraphics[height=2.25cm, width=1.5cm]{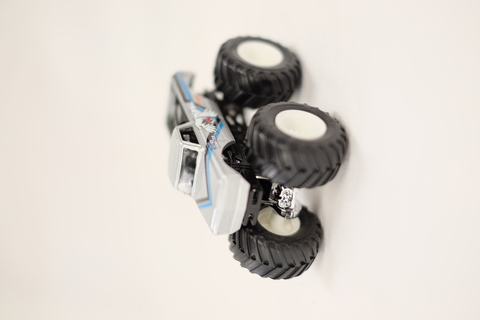} \hspace{1pt}
	\includegraphics[height=2.25cm, width=1.5cm]{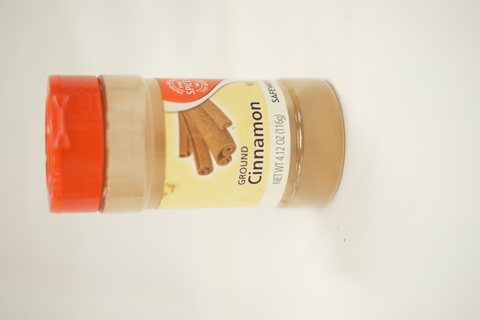} \hspace{1pt}
	\includegraphics[height=2.25cm, width=1.5cm]{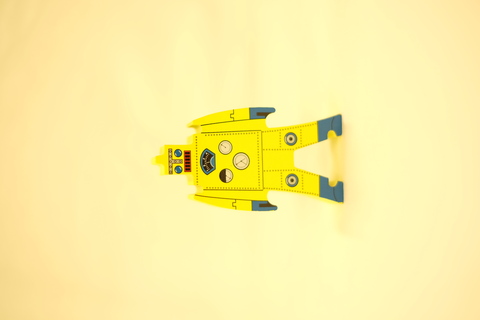} \hspace{1pt}
	\includegraphics[height=2.25cm, width=1.5cm]{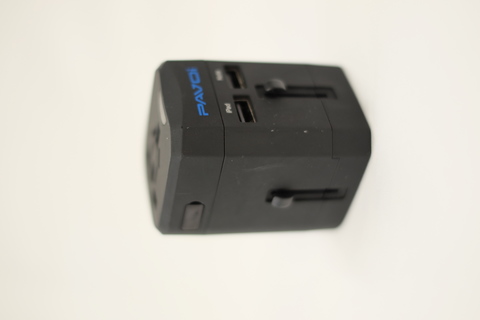} \hspace{1pt}
	\includegraphics[height=2.25cm, width=1.5cm]{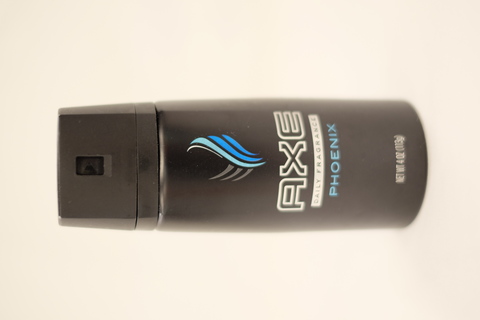} \hspace{1pt}
	\includegraphics[height=2.25cm, width=1.5cm]{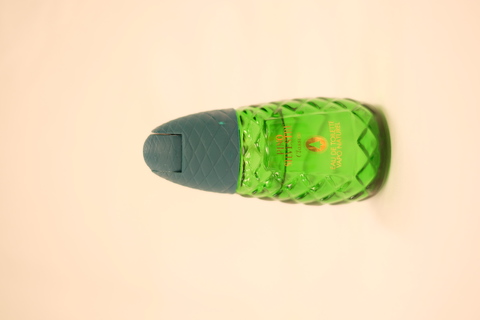} \hspace{1pt}
	\includegraphics[height=2.25cm, width=1.5cm]{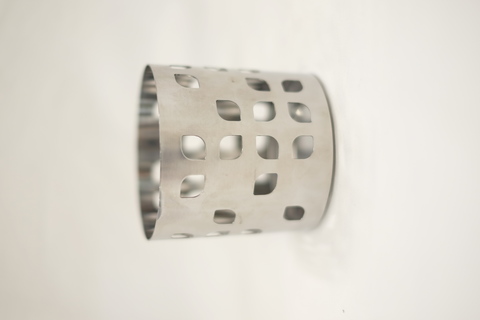}
	\\ \vspace{5pt}
	\includegraphics[height=2.25cm, width=1.5cm]{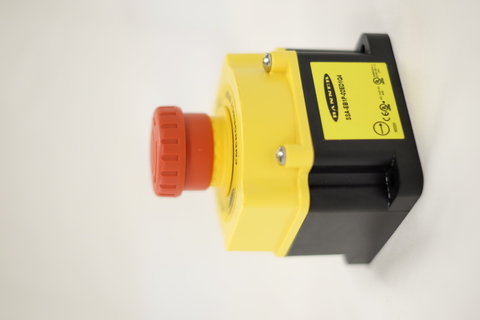} \hspace{1pt}
	\includegraphics[height=2.25cm, width=1.5cm]{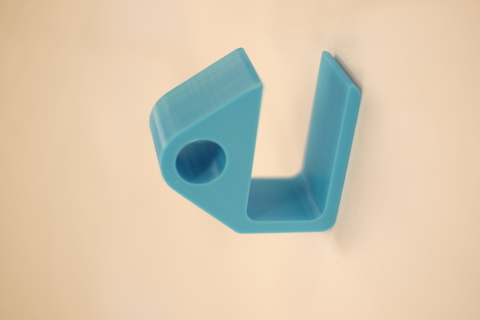} \hspace{1pt}
	\includegraphics[height=2.25cm, width=1.5cm]{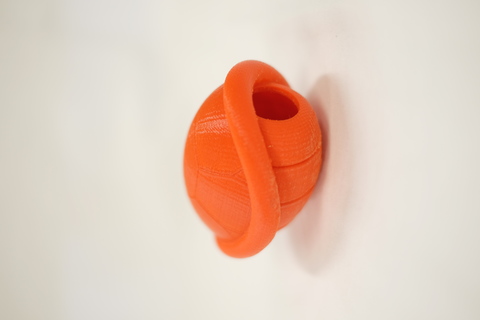} \hspace{1pt}
	\includegraphics[height=2.25cm, width=1.5cm]{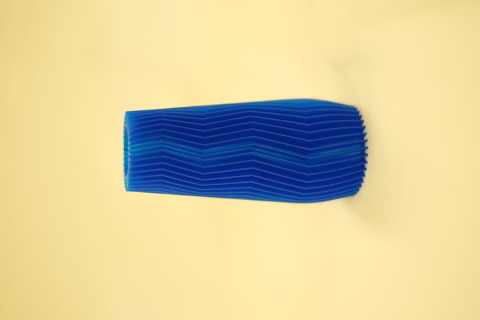} \hspace{1pt}
	\includegraphics[height=2.25cm, width=1.5cm]{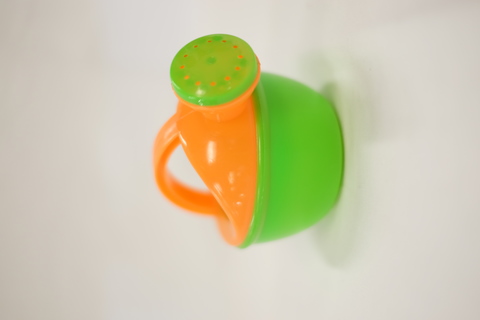} \hspace{1pt}
	\includegraphics[height=2.25cm, width=1.5cm]{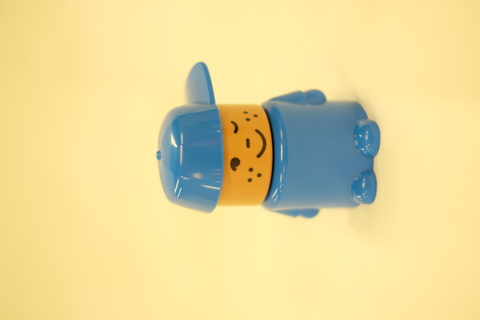} \hspace{1pt}
	\includegraphics[height=2.25cm, width=1.5cm]{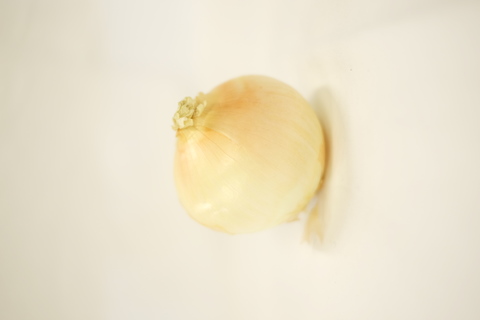} \hspace{1pt}
	\includegraphics[height=2.25cm, width=1.5cm]{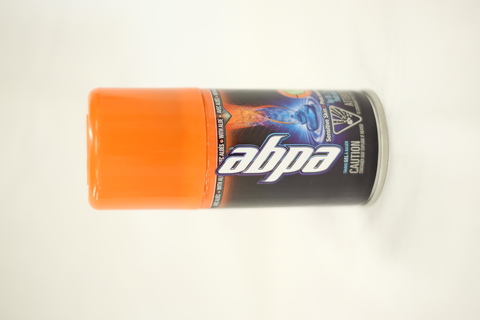}
	\\ \vspace{5pt}
	\includegraphics[height=2.25cm, width=1.5cm]{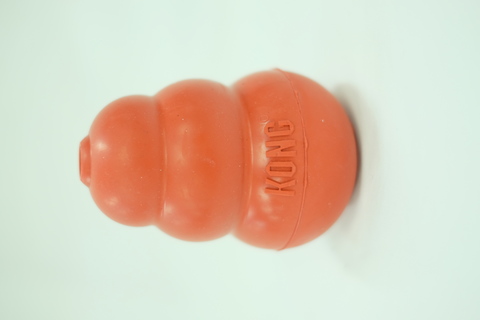} \hspace{1pt}
	\includegraphics[height=2.25cm, width=1.5cm]{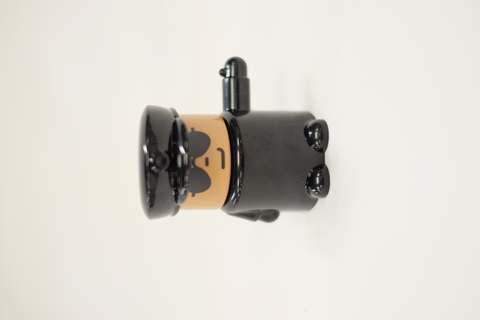} \hspace{1pt}
	\includegraphics[height=2.25cm, width=1.5cm]{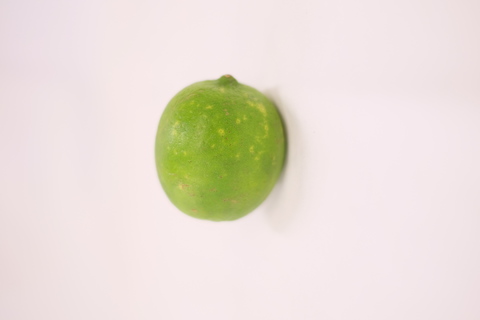} \hspace{1pt}
	\includegraphics[height=2.25cm, width=1.5cm]{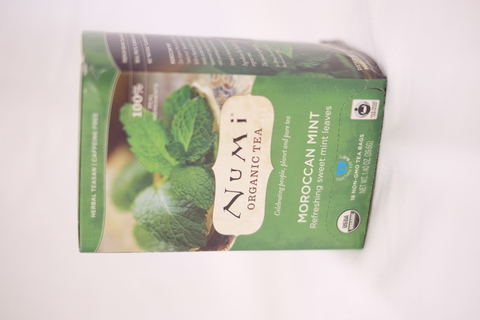} \hspace{1pt}
	\includegraphics[height=2.25cm, width=1.5cm]{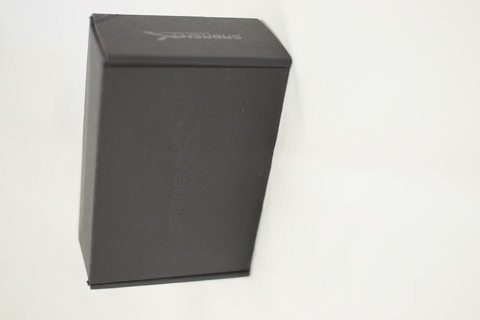} \hspace{1pt}
	\includegraphics[height=2.25cm, width=1.5cm]{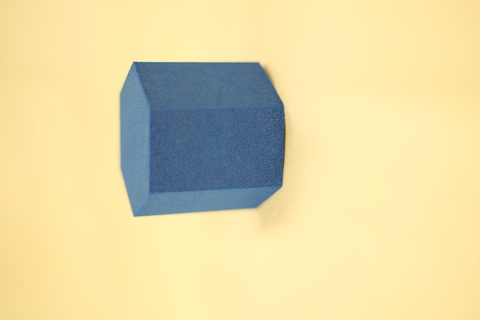} \hspace{1pt}
	\includegraphics[height=2.25cm, width=1.5cm]{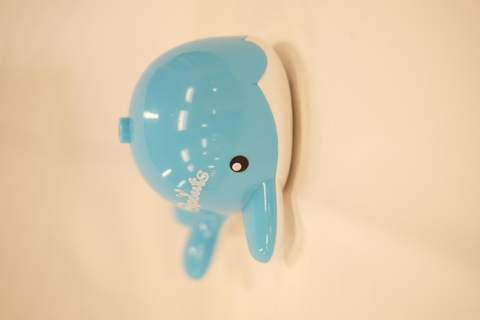} \hspace{1pt}
	\includegraphics[height=2.25cm, width=1.5cm]{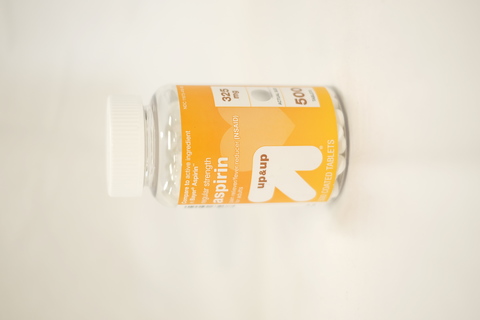}
	\\ \vspace{5pt}
	\includegraphics[height=2.25cm, width=1.5cm]{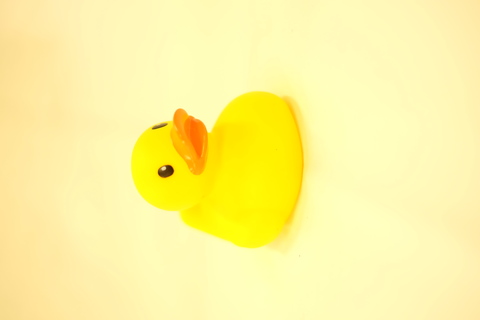} \hspace{1pt}
	\includegraphics[height=2.25cm, width=1.5cm]{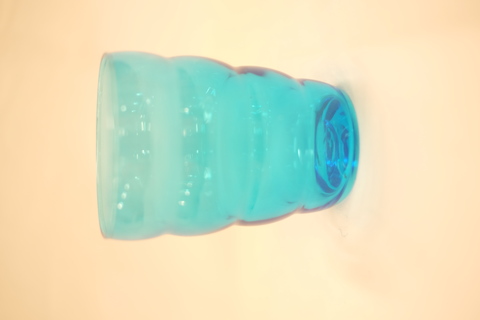} \hspace{1pt}
	\includegraphics[height=2.25cm, width=1.5cm]{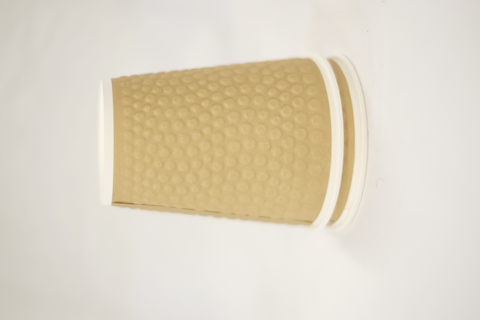} \hspace{1pt}
	\includegraphics[height=2.25cm, width=1.5cm]{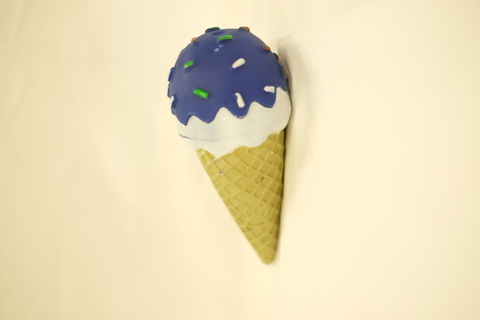} \hspace{1pt}
	\includegraphics[height=2.25cm, width=1.5cm]{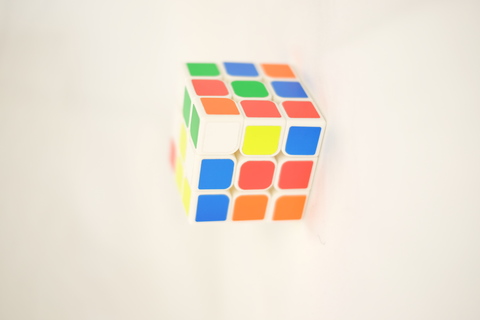} \hspace{1pt}
	\includegraphics[height=2.25cm, width=1.5cm]{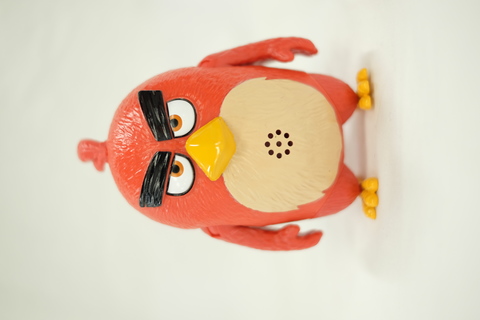} \hspace{1pt}
	\includegraphics[height=2.25cm, width=1.5cm]{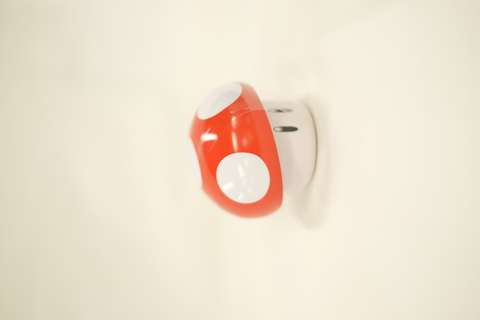} \hspace{1pt}
	\includegraphics[height=2.25cm, width=1.5cm]{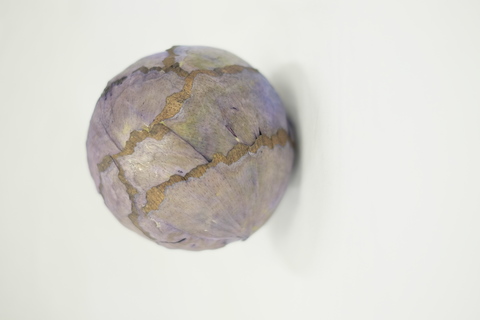}
	\caption{Examples of training objects. Overall, \nCollectedObj{} objects were used to collect \exactndatagrasping{} grasps.}
	\label{fig:training_objects}
	\vspace{-10pt}
\end{figure}

% Data labeling
The labels for this data were automatically generated using a deep neural network classifier trained to detect contacts using the raw \gelsight{} images measured at $T_c$.
%\footnote{For each finger we separately classified if the finger was in contact, and if both fingers were classified as in contact, the grasp would be considered successful.}
We performed additional manual labeling on small set of the collected data for which the automatic classification was borderline ambiguous, or in the rare cases when a visual inspection would indicate a wrong label.
Overall, we collected \exactndatagrasping{} grasping trials from \nCollectedObj{} unique objects, some of which are shown in \fig{fig:training_objects}.

%

%===============================================================================

\section{Experimental Results}
\label{sec:result}

In this section, we compare the predictive performance of our multi-modal visuo-tactile model against other models and other sensory modalities.
We evaluate the models by 1) comparing their accuracy at predicting grasp outcome on unseen test data  2) evaluating their performance at actually choosing successful grasps in a real-world robotic experiment.
The dataset used, code and videos of grasping trials are available on the website \website{}.

\subsection{Comparison on Outcome Prediction Accuracy}
\label{sec:4:classifier}
	How does a model with tactile sensing compare to a model trained on other sensory inputs?  
	Does combining multiple modalities help predict whether a grasp will be successful? 
	To address these questions, we explored several variations of our model, for each one measuring the ability to predict whether a grasp will be successful.

	\paragraph{Evaluation procedure} 
	Given the dataset collected as described in \sec{sec:setting}, we divided our dataset into a training and test set, such that each object only appears in one set. 	Using this data, we measured the predictive power of each model (see \tab{tab:accuracy}). We repeated these experiments over three random splits of the objects, and averaged the results.

	\paragraph{Multi-modal models} 
	First, we studied a {\em tactile} model which operates on GelSight images before and at the moment of the grasp from two sensors.  
	We then compared its performance to two visual models: one that operated on color images before and during the grasp, and one that used depth images (both were recorded from the same viewpoint using the RGB-D sensor). 
	We also explored combinations of modalities.  
	In particular, our {\em vision + tactile} model fuses the visual and tactile models (we chose RGB images for this, as a representative example of a visual modality).
	We also tried providing the location and orientation of the gripper as additional information to the visual network, which can function as an additional localization cue.  
	For this, we represent the gripper parameters using a three-layer fully connected network.
	We also considered hand-crafted tactile features. 
	Inspired by \citet{Dong2017}, which used pixel intensity as a measure of indentation, we computed the average absolute difference between pair of GelSight images captured before and during the grasp.  
	Note that this feature conveys both the surface area of the indentation, which is approximately the number of pixels that differ in the two images, and the amount of force (the magnitude of the intensity).  
	We included one such feature for both touch sensors, averaging over color channels.  
	We then trained a linear SVM on these features, which in \tab{tab:accuracy} we call the {\em Indentation} model.  
	Finally, since gel-based sensors contain large amounts of variation in their outputs, we trained versions of our model with only one fingertip sensor.  
	This variation may be from inter-sensor variation, which is caused by differences in the appearance of the silicone gel between sensors, or from intra-sensor variation, caused by wear-and-tear that the gel undergoes as it repeatedly grasps objects.

	\paragraph{Analysis} 
	We see in \tab{tab:accuracy} that the tactile model significantly outperformed the visual model. Furthermore, we see an improvement from combining visual and tactile information, with the multi-modal visual-tactile model performing the best of all models in the evaluation. 
	This indicates that combining multiple complementary modalities can improve grasp outcome prediction.  
	\begin{wraptable}{r}{0.45\linewidth}
	\centering
	\begin{tabular}{l|c}
	Method & Test acc. (\%) \\
	\thickhline
	Tactile + vision &$\mathbf{77.8 \pm 0.3\%}$ \\
	Vision only & $68.8 \pm 1.0\% $\\
    Vision + Gripper pose & $68.8 \pm 1.3\%$ \\
  	Depth & $73.2 \pm 0.7\%$\\
	Tactile (Both)& $75.6 \pm 0.8\% $\\
	Tactile (GelSight L) & $75.3 \pm 1.4\%$\\ % A
	Tactile (GelSight R) & $73.8 \pm 1.7\%$\\ % B
	Indentation features & $72.7 \pm 0.8\%$\\
    Chance & $61.8 \pm 1.9\%$\\
	\thickhline
	\end{tabular}
	  \caption{Classification accuracy (report mean and standard error) for models trained with different modalities as input. The models trained with tactile sensing achieve better accuracy than purely visual models.  Furthermore, our multi-modal model achieves the best overall results.}
	  \label{tab:accuracy}
% 	  \vspace{-5pt}
	\end{wraptable}
		The strong performance of the hand-crafted features is in part explained by the small size of the dataset, which limits the capabilities of large, expressive models. 
	However, since the principle aim of our experiments is to evaluate the relative importance of touch sensing in a combined visuo-tactile model, we used only the end-to-end trained models in the remainder of the experiments (\ie, the grasp performance evaluation in Section~\ref{sec:4:grasping}). 
	Since the end-to-end models integrate touch and visual sensing through the same type of convolutional network, they provide a fair setting for evaluating the benefits of each modality. 
	Further analysis of the relative performance of the hand-designed and learned features is left for future work, and is likely to be more feasible with larger datasets, though our current experiments suggest that end-to-end training does provide improved performance over manually designed features.
	
	The model that used both tactile sensors outperformed the models that used only one, but we also found variations in the predictive power of different \gelsight{} sensors.  
	This result may be due to the fact that the gel attached to the better-performing model was replaced fewer times during the data collection process, while the other sensor's gel was replaced more often due to tearing and slippage, thus introducing a factor another variability in the data and potentially making the learning process harder.

\subsection{Evaluation of Grasping Performance}
\label{sec:4:grasping}

	While evaluating model prediction accuracy on test data can give us a sense for the predictive power of each model and sensory modality, in practice we are more interested in the ability of the model to actually help us choose good grasps in the real world. 
	To evaluate this, we conducted a real-world grasping experiment using our robot setup and compared multiple models.
	To verify the generalization capabilities of the various approaches, we tested on \nGraspingObj{} new objects that were never seen in either the training or test set, which are shown in \tab{tab:grasping_success}.
	For each object, we repeated the experiment 10 times, and we choose grasps using the selection method described in \sec{sec:optimization}. 
	The models were retrained on all available data (\ie, both the training set and the test set used in \sec{sec:4:classifier}).
	As first baseline, we evaluated the manually engineered image-based grasping procedure used to autonomously collecting the data, as described in \sec{sec:setting}.
	This baseline made use of depth-sensors to fit a cylinder around the object and subsequently randomized the grasp pose and the force applied.
	Since we used this method for autonomously collecting data, it was quite heavily fine-tuned to perform well.
	The other two methods in our evaluation were the end-to-end trained model that used only vision, and the multi-modal model that combined vision and tactile sensing.
	
	The experimental results, presented in \tab{tab:grasping_success}, indicate that the multi-modal tactile+vision model outperform the other approaches, with a 14\% improvement over grasp selection using vision alone.
	The success rates for individual objects, shown in \tab{tab:grasping_success}, indicate that the objects varied considerably in difficulty. 
	The computer mouse (object 1) in particular, was extremely difficult to grasp with the baseline and visual model unable to pick it up reliably, while the vision + tactile model picked it up $60\%$ of the time.
	We also observed that, in several cases, the purely visual model attempted to lift objects when one of the fingers was not making contact, while the visuo-tactile model never exhibited this problem, since such empty grasps are easily recognized with tactile sensing.
	Overall, the experimental results suggest that learning an end-to-end multi-modal model that makes use of both vision and rich tactile sensors is beneficial both in terms of predictive accuracy, and when employed for actual grasping on a real robotic system.

	\newcommand{\heightfig}[0]{1.6cm}
		\begin{table}[t]
		\includegraphics[height=\heightfig]{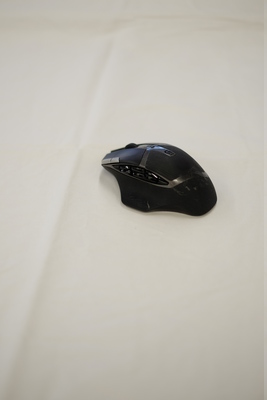} \hfill
		\includegraphics[height=\heightfig]{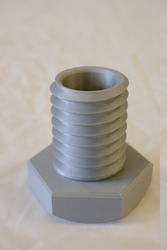} \hfill
		\includegraphics[height=\heightfig]{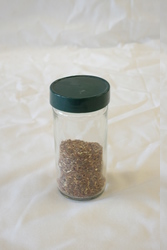} \hfill
		\includegraphics[height=\heightfig]{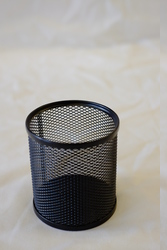} \hfill
		\includegraphics[height=\heightfig]{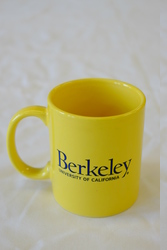} \hfill
		\includegraphics[height=\heightfig]{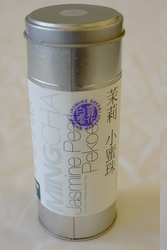} \hfill
		\includegraphics[height=\heightfig]{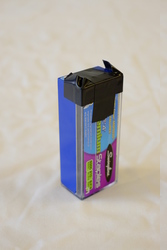} \hfill
		\includegraphics[height=\heightfig]{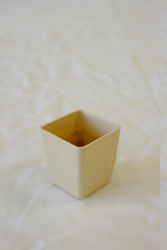} \hfill
		\includegraphics[height=\heightfig]{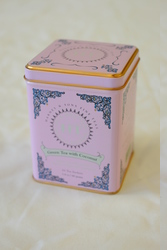} \hfill
		\includegraphics[height=\heightfig]{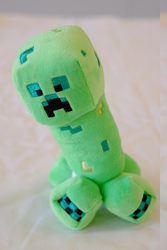} \hfill
		\includegraphics[height=\heightfig]{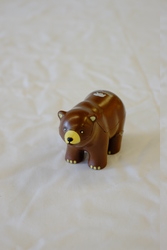} \hfill
		\includegraphics[height=\heightfig]{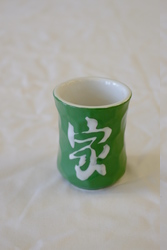}\\
		\vspace{-8pt}
	\centering
	\resizebox{\textwidth}{!}{%
	  \begin{tabular}{ | l | c | c | c | c | c | c | c | c | c | c|c|c|c|}
	    \hline
	    \multirow{2}{*}{Grasping Scheme}  & \multicolumn{12}{c|}{Grasping Successes (\% out of 10 trial)} & \multirow{2}{*}{TOT} \\ \cline{2-13}
	    & obj. 1 & obj. 2 & obj. 3 & obj. 4 & obj. 5 & obj. 6 & obj. 7 & obj. 8 & obj. 9& obj. 10& obj. 11& obj. 12&\\ \hline
	    Data Collection  & 0\% &	60\%&	10\%&	100\%	&10\%&	30\%&	20\% & 50\% & 30\% & 30\% &80\% & 100\% & 43\%\\ \hline
	    Vision only & 10\% & 100\% & 50\% & 100\% & 90\% & 100\% & 100\% & 80\% & 70\% &80\%& 80\% &100\% &\successVis\%\\ \hline
	    Tactile + vision & 60\% & 100\%	& 100\%	& 100\%	&70\%	&100\%	&100\% & 100\% & 100\% &100\% & 100\% & 100\% &\textbf{\successVisTac\%}
 \\ \hline\end{tabular}
	  } \vspace{1.5mm}
	  \caption{Grasping success rates for different grasp outcome prediction models. The evaluation is performed on \nGraspingObj{} objects not seen during training, which significantly differ in shape, dimension, weight, color, and material from the training objects. The ``data collection'' row shows the baseline results obtained with the detection method used for data collection. Incorporating tactile sensing (last row) yields the best results, with large improvements on several of the objects.}
	  \label{tab:grasping_success}
	\end{table}
	%

%===============================================================================

\section{Discussion and Future Work}
\label{sec:conclusion}

	% Recap paper and contributions
In this paper, we aimed to answer the following question: does touch sensing help predict grasp outcome, compared to purely visual perception?
To study this question, we proposed an end-to-end approach for predicting grasp outcome using raw visual and tactile inputs, using a tactile sensor that provides detailed information about contacts, forces, and compliance.
Our end-to-end approach does not require any characterization of the tactile sensors, nor a model of the robot or object.
As a result, our method requires minimal engineering of the actual grasping system, instead learning suitable visuo-tactile representations from data.
We autonomously collected more then \approxndatagrasping{} grasps and used them to train multiple deep neural network model for predicting grasp outcomes from different input modalities.
Our results indicate that the visuo-tactile multi-modal model substantially improves our ability to predict grasp outcomes compared to models that are based on only a single sensorial modality (\eg, vision).
To further validate this result, we performed a real-world evaluation of the different models in an active grasp selection.
Our experimental results demonstrate that the visuo-tactile multi-modal model outperform the vision only model, by achieving on \nGraspingObj{} previously unseen objects a grasp success rate of \successVisTac{}\%, compared to \successVis\% of the vision only model.

% Limitations
Although the proposed approach allowed us to study the importance of tactile sensing for robotic grasping, this method by itself is not necessarily an ideal solution for practical robotic grasping.
The learned model provides a rejection mechanism that evaluates the outcome of a grasp after the grasp is actually performed (but before liftoff).
As such, based on the quality of the proposing mechanism, convergence to an acceptable grasp might take an arbitrarily high number of grasp attempts.
In our experiments, we used a basic random search to propose new grasp location, which was sufficient to address the main experimental hypothesis of our work.
Although in most of the experiments a good solution would typically be accepted within 2 or 3 grasps, some trials took substantially longer, with one trial requiring 45 grasps before lifting the object successfully.
Moreover, our approach does not explicitly consider any post-liftoff event, such as slipping, hence not making full use of the interactive nature of tactile information for grasping.
Future work should aim to overcome these two limitations by proposing visuo-tactile solutions that are practical for real-world robot grasping. 

% Future work
The results obtained demonstrate the importance of tactile sensing for real-world robot grasping, and the effectiveness of deep neural network models to learn directly from raw visuo-tactile inputs.
Now that the importance of tactile sensing in robot grasping has been demonstrated, a new question arises: How to efficiently integrate tactile sensors to select successful grasp configurations?

%===============================================================================

% The maximum paper length is 8 pages excluding references and acknowledgements, and 10 pages including references and acknowledgements

\acknowledgments{We thank Chris Myers, Dan Chapman and the CITRIS invention lab for their support with 3D printing, and Siyuan Dong for the technical support with the \gelsight{}s. This research was supported by Berkeley DeepDrive, Honda, the Office of Naval Research through a Young Investigator Award, Toyota Research institute, Lincoln Laboratory,  DARPA (grant BAA-15-58), and the National Science Foundation. We also thank NVIDIA for equipment donations.
}

%===============================================================================

\bibliography{paper-tactile}  % .bib

\begin{thebibliography}{36}
\providecommand{\natexlab}[1]{#1}
\providecommand{\url}[1]{\texttt{#1}}
\expandafter\ifx\csname urlstyle\endcsname\relax
  \providecommand{\doi}[1]{doi: #1}\else
  \providecommand{\doi}{doi: \begingroup \urlstyle{rm}\Url}\fi

\bibitem[Johansson and Flanagan(2009)]{Johansson2009}
R.~S. Johansson and J.~R. Flanagan.
\newblock Coding and use of tactile signals from the fingertips in object manipulation tasks.
\newblock \emph{Nature Reviews Neuroscience}, 10\penalty0 (5):\penalty0 345--359, 2009.

\bibitem[Howe(1993)]{Howe1993}
R.~D. Howe.
\newblock Tactile sensing and control of robotic manipulation.
\newblock \emph{Advanced Robotics}, 8\penalty0 (3):\penalty0 245--261, 1993.
\newblock \doi{10.1163/156855394X00356}.

\bibitem[Bicchi et~al.(1988)Bicchi, Bergamasco, Dario, and Fiorillo]{Bicchi1988}
A.~Bicchi, M.~Bergamasco, P.~Dario, and A.~Fiorillo.
\newblock Integrated tactile sensing for gripper fingers.
\newblock In \emph{Proc. Int. Conf. on Robot Vision and Sensory Control}, 1988.

\bibitem[Romano et~al.(2011)Romano, Hsiao, Niemeyer, Chitta, and Kuchenbecker]{Romano2011}
J.~M. Romano, K.~Hsiao, G.~Niemeyer, S.~Chitta, and K.~J. Kuchenbecker.
\newblock Human-inspired robotic grasp control with tactile sensing.
\newblock \emph{IEEE Transactions on Robotics}, 27\penalty0 (6):\penalty0 1067--1079, 2011.
\newblock \doi{10.1109/TRO.2011.2162271}.

\bibitem[Veiga et~al.(2015)Veiga, van Hoof, Peters, and Hermans]{Veiga2015}
F.~Veiga, H.~van Hoof, J.~Peters, and T.~Hermans.
\newblock Stabilizing novel objects by learning to predict tactile slip.
\newblock In \emph{IEEE/RSJ Conference on Intelligent Robots and Systems (IROS)}, 2015.
\newblock \doi{10.1109/IROS.2015.7354090}.

\bibitem[Stachowsky et~al.(2016)Stachowsky, Hummel, Moussa, and Abdullah]{Stachowsky2016}
M.~Stachowsky, T.~Hummel, M.~Moussa, and H.~A. Abdullah.
\newblock A slip detection and correction strategy for precision robot grasping.
\newblock \emph{IEEE/ASME Transactions on Mechatronics}, 21\penalty0 (5):\penalty0 2214--2226, 2016.

\bibitem[Lenz et~al.(2015)Lenz, Lee, and Saxena]{Lenz2015}
I.~Lenz, H.~Lee, and A.~Saxena.
\newblock Deep learning for detecting robotic grasps.
\newblock \emph{The International Journal of Robotics Research}, 34\penalty0 (4-5):\penalty0 705--724, 2015.
\newblock \doi{10.1177/0278364914549607}.

\bibitem[Pinto and Gupta(2016)]{Pinto2016}
L.~Pinto and A.~Gupta.
\newblock Supersizing self-supervision: Learning to grasp from 50k tries and 700 robot hours.
\newblock In \emph{IEEE International Conference on Robotics and Automation (ICRA)}, pages 3406--3413, 2016.
\newblock \doi{10.1109/ICRA.2016.7487517}.

\bibitem[Levine et~al.(2016)Levine, Pastor, Krizhevsky, Ibarz, and Quillen]{Levine2016}
S.~Levine, P.~Pastor, A.~Krizhevsky, J.~Ibarz, and D.~Quillen.
\newblock Learning hand-eye coordination for robotic grasping with deep learning and large-scale data collection.
\newblock \emph{The International Journal of Robotics Research}, 2016.
\newblock \doi{10.1177/0278364917710318}.

\bibitem[Dong et~al.(2017)Dong, Yuan, and Adelson]{Dong2017}
S.~Dong, W.~Yuan, and E.~Adelson.
\newblock Improved gelsight tactile sensor for measuring geometry and slip.
\newblock In \emph{IEEE/RSJ International Conference on Intelligent Robots and Systems (IROS)}, 2017.

\bibitem[Rosales et~al.(2011)Rosales, Porta, and Ros]{Rosales2011}
C.~Rosales, J.~M. Porta, and L.~Ros.
\newblock Global optimization of robotic grasps.
\newblock \emph{Proceedings of Robotics: Science and Systems VII}, 2011.

\bibitem[Shimoga(1996)]{Shimoga1996}
K.~B. Shimoga.
\newblock Robot grasp synthesis algorithms: A survey.
\newblock \emph{The International Journal of Robotics Research}, 15\penalty0 (3):\penalty0 230--266, 1996.

\bibitem[Goldfeder and Allen(2011)]{Goldfeder2011}
C.~Goldfeder and P.~K. Allen.
\newblock Data-driven grasping.
\newblock \emph{Autonomous Robots}, 31\penalty0 (1):\penalty0 1--20, 2011.
\newblock ISSN 1573-7527.
\newblock \doi{10.1007/s10514-011-9228-1}.

\bibitem[Rodriguez et~al.(2012)Rodriguez, Mason, and Ferry]{Rodriguez2012}
A.~Rodriguez, M.~T. Mason, and S.~Ferry.
\newblock From caging to grasping.
\newblock \emph{The International Journal of Robotics Research}, 31\penalty0 (7):\penalty0 886--900, 2012.
\newblock \doi{10.1177/0278364912442972}.

\bibitem[Kamon et~al.(1996)Kamon, Flash, and Edelman]{Kamon1996}
I.~Kamon, T.~Flash, and S.~Edelman.
\newblock Learning to grasp using visual information.
\newblock In \emph{IEEE International Conference on Robotics and Automation (ICRA)}, volume~3, pages 2470--2476, 1996.
\newblock \doi{10.1109/ROBOT.1996.506534}.

\bibitem[Saxena et~al.(2008)Saxena, Driemeyer, and Ng]{Saxena2008}
A.~Saxena, J.~Driemeyer, and A.~Y. Ng.
\newblock Robotic grasping of novel objects using vision.
\newblock \emph{The International Journal of Robotics Research}, 27\penalty0 (2):\penalty0 157--173, 2008.
\newblock \doi{10.1177/0278364907087172}.

\bibitem[Kappler et~al.(2015)Kappler, Bohg, and Schaal]{Kappler2015}
D.~Kappler, J.~Bohg, and S.~Schaal.
\newblock Leveraging big data for grasp planning.
\newblock In \emph{IEEE International Conference on Robotics and Automation (ICRA)}, pages 4304--4311, 2015.

\bibitem[Johns et~al.(2016)Johns, Leutenegger, and Davison]{Johns2016}
E.~Johns, S.~Leutenegger, and A.~J. Davison.
\newblock Deep learning a grasp function for grasping under gripper pose uncertainty.
\newblock In \emph{IEEE/RSJ International Conference on Intelligent Robots and Systems (IROS)}, pages 4461--4468, 2016.
\newblock \doi{10.1109/IROS.2016.7759657}.

\bibitem[Mahler et~al.(2016)Mahler, Pokorny, Hou, Roderick, Laskey, Aubry, Kohlhoff, Kr{\"o}ger, Kuffner, and Goldberg]{Mahler2016}
J.~Mahler, F.~T. Pokorny, B.~Hou, M.~Roderick, M.~Laskey, M.~Aubry, K.~Kohlhoff, T.~Kr{\"o}ger, J.~Kuffner, and K.~Goldberg.
\newblock Dex-net 1.0: A cloud-based network of 3d objects for robust grasp planning using a multi-armed bandit model with correlated rewards.
\newblock In \emph{IEEE International Conference on Robotics and Automation (ICRA)}, pages 1957--1964, 2016.

\bibitem[Viereck et~al.(2017)Viereck, Pas, Saenko, and Platt]{Viereck2017}
U.~Viereck, A.~t. Pas, K.~Saenko, and R.~Platt.
\newblock Learning a visuomotor controller for real world robotic grasping using easily simulated depth images.
\newblock \emph{arXiv preprint arXiv:1706.04652}, 2017.

\bibitem[Gualtieri et~al.(2016)Gualtieri, ten Pas, Saenko, and Platt]{Gualtieri2016}
M.~Gualtieri, A.~ten Pas, K.~Saenko, and R.~Platt.
\newblock High precision grasp pose detection in dense clutter.
\newblock In \emph{IEEE/RSJ International Conference on Intelligent Robots and Systems (IROS)}, pages 598--605, 2016.

\bibitem[Bohg et~al.(2014)Bohg, Morales, Asfour, and Kragic]{Bohg2014}
J.~Bohg, A.~Morales, T.~Asfour, and D.~Kragic.
\newblock Data-driven grasp synthesis -- a survey.
\newblock \emph{IEEE Transactions on Robotics}, 30\penalty0 (2):\penalty0 289--309, 2014.
\newblock \doi{10.1109/TRO.2013.2289018}.

\bibitem[Yousef et~al.(2011)Yousef, Boukallel, and Althoefer]{Yousef2011}
H.~Yousef, M.~Boukallel, and K.~Althoefer.
\newblock Tactile sensing for dexterous in-hand manipulation in robotics—a review.
\newblock \emph{Sensors and Actuators A: physical}, 167\penalty0 (2):\penalty0 171--187, 2011.

\bibitem[Bekiroglu et~al.(2011)Bekiroglu, Laaksonen, Jorgensen, Kyrki, and Kragic]{Bekiroglu2011}
Y.~Bekiroglu, J.~Laaksonen, J.~A. Jorgensen, V.~Kyrki, and D.~Kragic.
\newblock Assessing grasp stability based on learning and haptic data.
\newblock \emph{IEEE Transactions on Robotics}, 27\penalty0 (3):\penalty0 616--629, 2011.

\bibitem[Schill et~al.(2012)Schill, Laaksonen, Przybylski, Kyrki, Asfour, and Dillmann]{Schill2012}
J.~Schill, J.~Laaksonen, M.~Przybylski, V.~Kyrki, T.~Asfour, and R.~Dillmann.
\newblock Learning continuous grasp stability for a humanoid robot hand based on tactile sensing.
\newblock In \emph{IEEE RAS \& EMBS International Conference on Biomedical Robotics and Biomechatronics (BioRob)}, pages 1901--1906. IEEE, 2012.

\bibitem[Li et~al.(2014)Li, Bekiroglu, Kragic, and Billard]{Li2014a}
M.~Li, Y.~Bekiroglu, D.~Kragic, and A.~Billard.
\newblock Learning of grasp adaptation through experience and tactile sensing.
\newblock In \emph{IEEE/RSJ International Conference on Intelligent Robots and Systems (IROS)}, pages 3339--3346, 2014.

\bibitem[Allen et~al.(1999)Allen, Miller, Oh, and Leibowitz]{Allen1999}
P.~K. Allen, A.~T. Miller, P.~Y. Oh, and B.~S. Leibowitz.
\newblock Integration of vision, force and tactile sensing for grasping.
\newblock 1999.

\bibitem[Bekiroglu(2012)]{Bekiroglu2012}
Y.~Bekiroglu.
\newblock \emph{Learning to Assess Grasp Stability from Vision, Touch and Proprioception}.
\newblock PhD thesis, KTH Royal Institute of Technology, 2012.

\bibitem[Jara et~al.(2014)Jara, Pomares, Candelas, and Torres]{Jara2014}
C.~A. Jara, J.~Pomares, F.~A. Candelas, and F.~Torres.
\newblock Control framework for dexterous manipulation using dynamic visual servoing and tactile sensors’ feedback.
\newblock \emph{Sensors}, 14\penalty0 (1):\penalty0 1787--1804, 2014.
\newblock \doi{10.3390/s140101787}.

\bibitem[Johnson and Adelson(2009)]{Johnson2009}
M.~K. Johnson and E.~Adelson.
\newblock Retrographic sensing for the measurement of surface texture and shape.
\newblock In \emph{IEEE Conference on Computer Vision and Pattern Recognition (CVPR)}, pages 1070--1077, 2009.

\bibitem[Yuan et~al.(2017{\natexlab{a}})Yuan, Zhu, Owens, Srinivasan, and Adelson]{Yuan2017}
W.~Yuan, C.~Zhu, A.~Owens, M.~A. Srinivasan, and E.~H. Adelson.
\newblock Shape-independent hardness estimation using deep learning and a gelsight tactile sensor.
\newblock \emph{arXiv preprint arXiv:1704.03955}, 2017{\natexlab{a}}.

\bibitem[Yuan et~al.(2017{\natexlab{b}})Yuan, Wang, Dong, and Adelson]{Yuan2017a}
W.~Yuan, S.~Wang, S.~Dong, and E.~Adelson.
\newblock Connecting look and feel: Associating the visual and tactile properties of physical materials.
\newblock \emph{arXiv preprint arXiv:1704.03822}, 2017{\natexlab{b}}.

\bibitem[Ngiam et~al.(2011)Ngiam, Khosla, Kim, Nam, Lee, and Ng]{Ngiam2011}
J.~Ngiam, A.~Khosla, M.~Kim, J.~Nam, H.~Lee, and A.~Y. Ng.
\newblock Multimodal deep learning.
\newblock In \emph{International Conference on Machine Learning (ICML)}, 2011.

\bibitem[He et~al.(2016)He, Zhang, Ren, and Sun]{he2016deep}
K.~He, X.~Zhang, S.~Ren, and J.~Sun.
\newblock Deep residual learning for image recognition.
\newblock In \emph{Proceedings of the IEEE conference on computer vision and pattern recognition}, pages 770--778, 2016.

\bibitem[Deng et~al.(2009)Deng, Dong, Socher, Li, Li, and Fei-Fei]{Deng2009}
J.~Deng, W.~Dong, R.~Socher, L.-J. Li, K.~Li, and L.~Fei-Fei.
\newblock Imagenet: A large-scale hierarchical image database.
\newblock In \emph{IEEE Conference on Computer Vision and Pattern Recognition (CVPR)}, pages 248--255, 2009.

\bibitem[Kingma and Ba(2014)]{Kingma2014}
D.~Kingma and J.~Ba.
\newblock Adam: A method for stochastic optimization.
\newblock \emph{arXiv preprint arXiv:1412.6980}, 2014.

\end{thebibliography}

\end{document}